%% file: main.tex
\documentclass[11pt]{article}

\PassOptionsToPackage{table}{xcolor}
\usepackage[preprint]{acl}
\usepackage{times}
\usepackage{latexsym}
\usepackage[T1]{fontenc}
\usepackage[utf8]{inputenc}
\usepackage{microtype}
\usepackage{graphicx}
\usepackage{booktabs}
\usepackage{multirow}
\usepackage{tabularx}
\usepackage{array}
\usepackage{amsmath}
\usepackage{amssymb}
\usepackage{amsthm}
\usepackage{mathtools}
\usepackage{algorithm}
\usepackage[noend]{algpseudocode}
\usepackage{enumitem}

\input{appendix_style}

\usepackage{url}
\usepackage{hyperref}
\usepackage{tikz}
\usetikzlibrary{arrows.meta,positioning,fit,calc,shapes.geometric}

\graphicspath{{figures/}}
\definecolor{smublue}{HTML}{0F4D92}
\definecolor{smugreen}{HTML}{4F9B5F}
\definecolor{smured}{HTML}{B64342}
\definecolor{smugray}{HTML}{767676}
\definecolor{groupopen}{HTML}{ECECFF}
\definecolor{groupclosed}{HTML}{FAEEE2}
\definecolor{groupmemory}{HTML}{EAF8E9}
\hypersetup{
  colorlinks=true,
  citecolor=smublue,
  linkcolor=smublue,
  urlcolor=smublue
}

\usepackage{etoolbox}
\makeatletter
\ifcsdef{@outputbox@appendfootnotes}{%
  \patchcmd{\@outputbox@appendfootnotes}
    {\unvbox\footins}
    {\pdfrunninglinkoff\unvbox\footins\pdfrunninglinkon}
    {}
    {\PackageError{memduet}{Could not protect footnote links}{Use pdfLaTeX with the supplied ACL style.}}%
}{%
  \patchcmd{\@makecol}
    {\unvbox\footins}
    {\pdfrunninglinkoff\unvbox\footins\pdfrunninglinkon}
    {}
    {\PackageError{memduet}{Could not protect footnote links}{Use pdfLaTeX with the supplied ACL style.}}%
}
\makeatother

\input{action_style}
\newcommand{\update}{\memaction{Update}}

\title{MemDuet: Learning Dual-Process Memory for Long-Context Reasoning}

\author{Jiangcheng Song\textsuperscript{1}, Haodong Jing\textsuperscript{1}, Yulin Mu\textsuperscript{2}, Lesong Tao\textsuperscript{1}, Jixin Wang\textsuperscript{1} \\
  \textbf{Jin Cui\textsuperscript{1}, Tianjing Ren\textsuperscript{3}, Runxuan Xie\textsuperscript{2}, Yongqiang Ma\textsuperscript{1}, Nanning Zheng\textsuperscript{1}\thanks{Corresponding author: \href{mailto:nnzheng@xjtu.edu.cn}{\texttt{nnzheng@xjtu.edu.cn}}.}} \\
  \textsuperscript{1}State Key Laboratory of Human-Machine Hybrid Augmented Intelligence, \\
  Institute of Artificial Intelligence and Robotics, \\
  Xi'an Jiaotong University \\
  \textsuperscript{2}Faculty of Electronic and Information Engineering, \\
  Xi'an Jiaotong University \\
  \textsuperscript{3}Imperial College London \\
  \texttt{enone@stu.xjtu.edu.cn, nnzheng@xjtu.edu.cn}}

\begin{document}
\maketitle

\begin{abstract}
Reasoning over infinitely long documents is becoming a reality with recurrent
memory agents. However, increasing context length exacerbates the trade-off
between reasoning performance and inference efficiency. Inspired by Kahneman's
account of fast and slow thinking, we propose \textbf{MemDuet}\footnote{Project page: \href{https://jiangchengsong.github.io/MemDuet/}{\nolinkurl{jiangchengsong.github.io/MemDuet/}}.}, a dual-process
memory framework that learns to coordinate a small language model for fast,
parallel evidence selection from raw context chunks (System 1) with a large
language model for memory-update verification and memory updates in document
order (System 2).
We train MemDuet in three stages: supervised fine-tuning of
System 1 for evidence selection, Group Relative Policy Optimization (GRPO) of
System 2 for verification and memory rewriting, and joint optimization of both
policies to coordinate selection and updating for long-context reasoning.
Experiments on diverse long-context reasoning benchmarks
(RULER-HQA, LongBench v2, and 2WikiMultiHopQA)
demonstrate a favorable accuracy--efficiency trade-off as context length grows.
Averaged across evaluation settings on the three datasets, MemDuet achieves an
end-to-end inference speedup of \textbf{9.74$\times$} and an
output-token reduction of \textbf{73.91\%} relative to RL-MemAgent.
\end{abstract}

\begin{figure}[!t]
\centering
\includegraphics[width=\columnwidth,trim=0 12bp 0 0,clip]{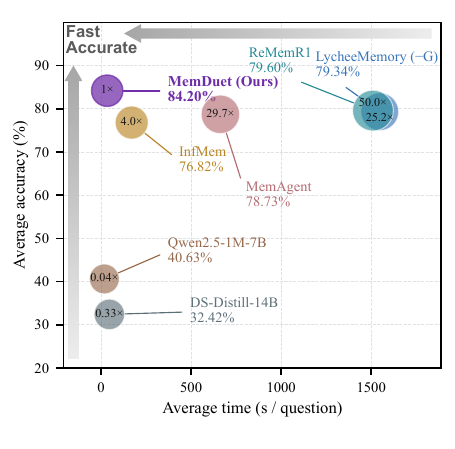}
\caption{Accuracy--efficiency trade-off on RULER-HQA. Accuracy, inference
time, and output tokens are equally averaged over the nine shared length
settings (7K--1.75M). Qwen and DS-Distill use
the available settings. Circle areas use a
square-root-compressed scale for mean output tokens per question; circle
labels give actual output-token ratios to MemDuet (1$\times$).}
\label{fig:intro-tradeoff}
\end{figure}

\section{Introduction}
\label{sec:intro}

Long-context reasoning is essential for large language models (LLMs) to
synthesize evidence across documents, understand lengthy reports, and resolve
repository-level dependencies in code
\citep{bai2024longbench,zhang2023repocoder}. Existing work primarily extends
long-context capability through larger context windows, sparse or linear
attention, and retrieval-augmented generation (RAG)
\citep{chen2023position,beltagy2020longformer,katharopoulos2020transformers,yuan2025nsa,lewis2020rag}.
However, these approaches remain bounded by the amount of context that can
be directly represented or retrieved, leaving reasoning over unbounded
documents unresolved.

MemAgent instead streams documents through a bounded recurrent memory,
enabling reasoning over infinitely long documents with a fixed active context
\citep{yu2026memagent}. Building on this paradigm, InfMem and GRU-Mem
learn when to retrieve, update, or stop, while LycheeMemory compresses each
chunk and invokes the reasoner only for selected memory blocks
\citep{wang2026infmem,sheng2026grumem,chen2026lycheememory}. These advances make
recurrent memory increasingly selective, but expose a common bottleneck:
control decisions either remain inside the sequential recurrent loop or
operate on representations in which the original evidence has already been
compressed. Consequently, as context grows, reasoning performance and
inference efficiency remain tightly coupled.

Human reading and memory suggest a different principle: memory formation begins
with selective attention. Readers first focus on goal-relevant information and
then integrate the attended evidence with prior knowledge into a coherent
memory representation \citep{craik1996divided,kintsch1988knowledge,aly2016attention}.
This process naturally maps onto the dual-process account of cognition:
\textbf{System 1} rapidly identifies potentially useful evidence, whereas
\textbf{System 2} deliberately assesses and consolidates it
\citep{kahneman2011thinking,evans2013dual}. Based on this insight, we propose
\textbf{MemDuet}, a dual-process recurrent memory framework. A small language
model (System 1) scans raw context in parallel to select candidate chunks for
memory updating. A large language model (System 2) verifies whether each
selected chunk warrants a memory update under the current memory, then
rewrites memory. By combining small-model selective attention over
raw chunks with large-model verification and memory updating, MemDuet balances
reasoning performance and inference efficiency as context grows.

A key challenge is how to train the two systems to cooperate. We develop a
progressive three-stage training strategy. First, we supervise System 1 to
identify context chunks that are likely to trigger useful memory updates. Next,
with System 1 fixed, we optimize System 2 using Group Relative Policy
Optimization (GRPO; \citealp{shao2024deepseekmath}), training it to verify
memory-update decisions and rewrite memory toward successful downstream reasoning.
Finally, we jointly optimize both systems with task-level rewards, allowing
evidence selection and memory updating to co-adapt.

We evaluate MemDuet on RULER-HQA, LongBench v2, and long-context
2WikiMultihopQA. Results show that MemDuet improves the
performance--efficiency trade-off over full-update recurrent memory agents,
scaling to 3.5M-token contexts while achieving higher reasoning accuracy
with substantially lower inference cost.

Our contributions are summarized as follows:
\begin{enumerate}[leftmargin=*,itemsep=1pt,topsep=3pt]
  \item We propose MemDuet, a cognitively inspired dual-process memory framework
  for long-context reasoning, which instantiates System 1 as fast,
  parallel evidence selection with a small language model and System 2
  as memory-update verification and memory rewriting with a large
  language model.
  \item We develop a three-stage training strategy that first establishes the
  selection and memory-updating capabilities of the two systems through SFT and
  GRPO, respectively, and then jointly optimizes their cooperation under a
  shared downstream reasoning objective.
  \item Extensive experiments on long-context reasoning benchmarks demonstrate
  that MemDuet achieves a favorable balance between reasoning performance and
  inference efficiency.
\end{enumerate}

\section{Related Work}
\label{sec:related}

\subsection{Recurrent Memory for Long-Context Reasoning}
Recurrent-memory methods process documents incrementally while maintaining a
bounded state, allowing language models to handle inputs far beyond their
context windows. MemAgent sequentially processes text chunks and repeatedly
rewrites a textual memory, using end-to-end reinforcement learning to retain
task-relevant information and discard distractors based on downstream rewards
\citep{yu2026memagent}. Building on this paradigm, InfMem and GRU-Mem learn
retrieval, memory updating, and early-stopping decisions through SFT-to-RL
training or behavior-specific rewards \citep{wang2026infmem,sheng2026grumem}.
LycheeMemory compresses individual chunks and selectively invokes the reasoner,
jointly optimizing its compressor and reasoner through end-to-end reinforcement
learning \citep{chen2026lycheememory}. These studies turn memory management from
a fixed procedure into a task-driven policy, but their control decisions remain
within the sequential loop or operate on already compressed evidence. MemDuet
instead scans raw chunks in parallel before recurrent updating and progressively
coordinates evidence selection with memory consolidation.

\input{sections/framework_figure_en}

\subsection{Cognitively Inspired Language Models and Memory}
Cognitive theories have informed both memory design and computational
allocation in language-model systems. CoALA organizes language agents around
modular memory, decision-making, and action components derived from cognitive
architectures \citep{sumers2024coala}. MemoryBank adopts an Ebbinghaus-inspired
mechanism for dynamically forgetting and reinforcing memories
\citep{zhong2024memorybank}, whereas HippoRAG draws on hippocampal indexing
theory to support associative knowledge retrieval
\citep{gutierrez2024hipporag}. For reasoning, System 2 Attention deliberately
rewrites the context to remove irrelevant information
\citep{weston2023system2attention}, while Dualformer learns switchable fast and
slow reasoning modes within a single model \citep{su2025dualformer}. These
studies primarily apply cognitive principles to memory organization, retrieval,
or the allocation of reasoning effort. MemDuet instead introduces dual-process
specialization into recurrent memory: System 1 selects raw candidate chunks in
parallel; System 2 verifies whether they warrant memory updates under the
current memory and consolidates their evidence.

\section{Methodology}
\label{sec:events}
\label{sec:method}

\subsection{Preliminaries}
\label{sec:preliminaries}

\paragraph{Problem setup.}
We consider query-conditioned reasoning over a long document $D$. Given a
query $q$, the goal is to generate an answer $y$ using evidence from $D$,
whose length may exceed the model's context window. We divide $D$ into $T$
chunks $(c_1,\ldots,c_T)$ in their original order, with $|c_i|\leq C$, where
$|\cdot|$ denotes token length and $C$ is the chunk-size budget. The query is
available throughout document processing.

\paragraph{Recurrent textual memory.}
Following MemAgent \citep{yu2026memagent}, an LLM carries information
across chunks through bounded textual memory, using the common interface
\begin{equation}
m'\sim\pi_\theta^{\mathrm{mem}}(\cdot\mid q,c_i,m),
\qquad |m'|\leq B,
\label{eq:prelim-memory-update}
\end{equation}
Here $m$ and $m'$ are the preceding and replacement memories, and $B$ is
the memory budget. Under the memory prompt, $\pi_\theta^{\mathrm{mem}}$
is the autoregressive product of System~2's next-token policy
$\pi_\theta^{\mathrm{gen}}$. The same model answers from $q$ and the
final memory under the answer prompt. Both $C$ and $B$ are independent
of document length.

\subsection{MemDuet Workflow}
\label{sec:inference-workflow}

\input{sections/workflow_memduet_en}

\input{sections/training_memduet_en}

\section{Experiments}
\label{sec:experiments}

\input{sections/integrated_experiments_en}

\section{Conclusion}

MemDuet combines memory-conditioned parallel evidence selection by a small
language model with large-model verification and document-order memory
updates. Three-stage training teaches both systems when to update memory
and how to consolidate evidence and coordinate decisions.
Experiments on three long-context reasoning benchmarks
show a favorable accuracy--efficiency trade-off, with greater efficiency
gains at longer contexts.

\section*{Limitations}
\label{sec:limitations}

Our evaluation is limited to three long-context benchmarks and mainly
Qwen-family backbones, so generalization to other domains and model families
remains uncertain. MemDuet is also less effective on shorter inputs, where
selective processing may discard useful evidence; moreover, evidence missed
by System~1 cannot be recovered by System~2. Memory updates in document order further
limit the practical benefits of parallel execution.

\section*{Ethical Considerations}
\label{sec:ethics}

MemDuet studies efficient reasoning over long documents using existing
benchmarks and pretrained language models. These resources can contain
inaccurate, biased, or sensitive material. Their use and redistribution
should respect the applicable licenses and access conditions. Benchmark
performance alone does not establish fairness, privacy protection, or
suitability for deployment.

Selective processing can omit relevant evidence or minority viewpoints,
while repeated memory updates can preserve unsupported claims and discard
qualifications. The verifier evaluates memory-update decisions and should
not be treated as an independent guarantee of factual correctness. For
consequential applications, users should retain access to original
documents, inspect supporting evidence, and require human review of
outputs. Additional evaluation across languages, domains, and population
groups is needed before such use.

Long documents, intermediate memories, and generated answers may expose
personal or confidential information. Deployments should minimize sensitive
data, restrict access to stored memories and logs, and define appropriate
retention and deletion procedures. Users should confirm authorization
before transmitting content to proprietary APIs and consider the
providers' data-handling policies.

Lower inference cost may make document analysis more accessible, but can
also facilitate large-scale surveillance or unauthorized profiling.
Applications should be limited to authorized, legitimate uses. Finally,
reduced latency and output-token counts do not directly quantify energy
consumption or carbon emissions; environmental benefits require dedicated
measurement that also accounts for training and total usage.

\bibliography{references}

\appendix
\setcounter{table}{0}
\renewcommand{\thetable}{A\arabic{table}}
\renewcommand{\theHtable}{A.\arabic{table}}

\input{sections/appendices_en}

\end{document}

%% file: appendix_style.tex
\usepackage{listings}
\usepackage{placeins}
\usepackage[most]{tcolorbox}

\definecolor{appendixink}{HTML}{484848}
\definecolor{appendixblue}{HTML}{315F9B}
\definecolor{appendixred}{HTML}{9A292B}
\definecolor{appendixpurple}{HTML}{855A8A}
\definecolor{appendixgreen}{HTML}{326747}

\lstdefinestyle{memoryprompt}{
  basicstyle=\fontsize{9}{10.5}\selectfont\ttfamily,
  columns=fullflexible,
  keepspaces=true,
  breaklines=true,
  breakatwhitespace=true,
  showstringspaces=false,
  frame=none,
  aboveskip=0pt,
  belowskip=0pt,
  moredelim=[s][\color{appendixink!75}]{<}{>},
  moredelim=[s][\color{appendixpurple}]{\{} {\}}
}

\tcbset{appendixpanel/.style={
  enhanced,
  boxrule=0.55pt,
  arc=1.5mm,
  outer arc=1.5mm,
  left=2mm,right=2mm,top=1.5mm,bottom=1.5mm,
  toptitle=1mm,bottomtitle=1mm,
  fonttitle=\small\bfseries,
  fontupper=\small,
  before skip=7pt,after skip=7pt
}}

\newcommand{\memorypromptpanel}[2]{%
  \tcbinputlisting{appendixpanel,breakable,
    colback=white,colframe=appendixink,
    colbacktitle=appendixink,coltitle=white,
    title={#1},listing only,
    listing options={style=memoryprompt},listing file={#2}}%
}

\newtcolorbox{memorycase}[1]{appendixpanel,breakable,
  fontupper=\fontsize{9.5}{11}\selectfont,
  colback=appendixblue!4!white,colframe=appendixblue,
  colbacktitle=appendixblue!13!white,coltitle=appendixblue!85!black,
  title={#1}}
\newtcolorbox{failurecase}[1]{appendixpanel,breakable,
  fontupper=\fontsize{9.5}{11}\selectfont,
  colback=appendixred!3!white,colframe=appendixred,
  colbacktitle=appendixred,coltitle=white,
  title={#1}}
\newcommand{\casestep}[2]{%
  \par\smallskip{\color{appendixblue}\bfseries #1}\par\nobreak #2%
}
\newcommand{\failurestep}[2]{%
  \par\smallskip{\color{appendixred}\bfseries #1}\par\nobreak #2%
}
\newcommand{\casecorrect}[1]{\textcolor{appendixgreen}{\textbf{#1}}}
\newcommand{\casewrong}[1]{\textcolor{appendixred}{\textbf{#1}}}
\newcommand{\caseboundary}{%
  \par\smallskip{\color{black!20}\hrule height 0.35pt}\smallskip
}

%% file: action_style.tex
\ifdefined\XeTeXversion
  \newfontfamily\memactionfont{texgyretermes-regular.otf}[Letters=SmallCaps]
\else
  \newcommand{\memactionfont}{\fontfamily{qtm}\fontshape{sc}\selectfont}
\fi
\DeclareRobustCommand{\memaction}[1]{%
  \ifmmode
    \text{\normalfont\memactionfont #1}%
  \else
    {\normalfont\memactionfont #1}%
  \fi
}
\DeclareRobustCommand{\skipaction}{\memaction{Skip}}

%% file: sections/framework_figure_en.tex
\begin{figure*}[!t]
\centering
\includegraphics[width=\textwidth]{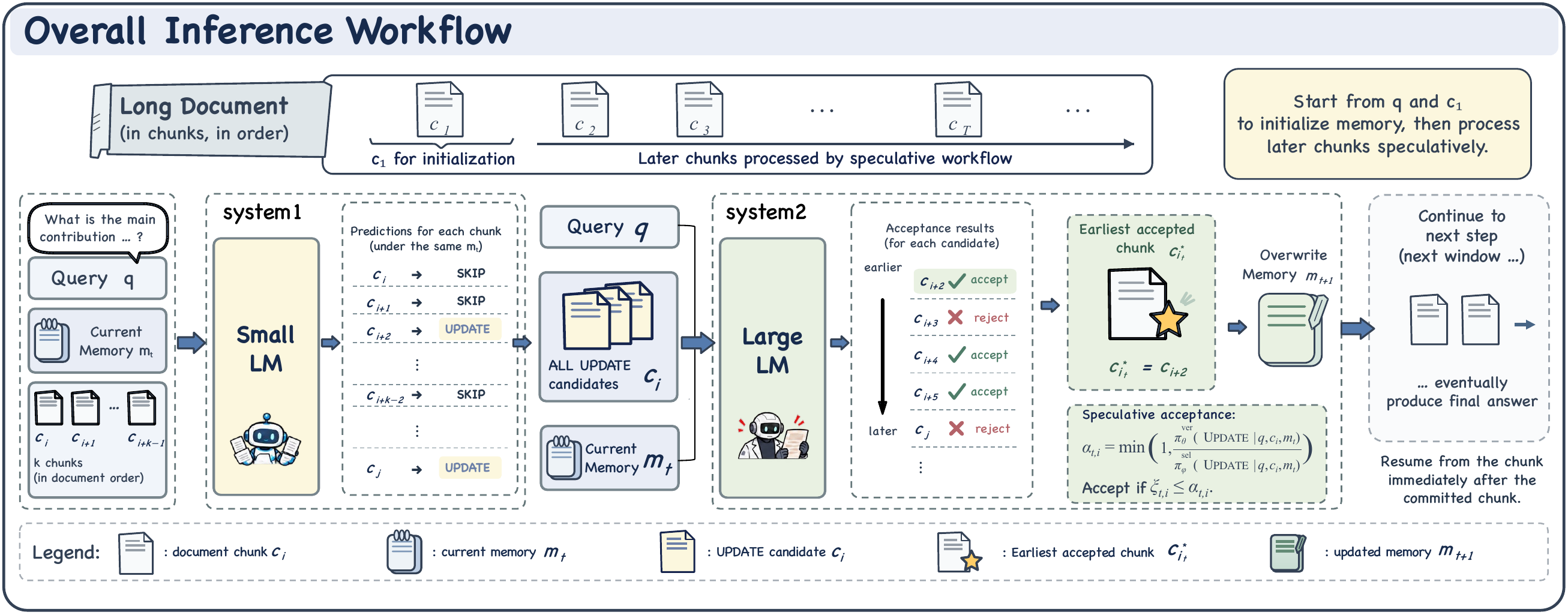}
\caption{MemDuet's overall inference workflow. System~1 predicts
\skipaction{}/\update{} in parallel for raw chunks under a shared memory
state. Candidate chunks retain their original text. System~2 verifies
whether a selected chunk warrants a memory update given the current memory,
then generates and commits memory from the first accepted chunk in document
order. Later chunks are reevaluated under the updated memory.}
\label{fig:framework}
\end{figure*}

%% file: sections/workflow_memduet_en.tex
MemDuet alternates parallel raw-chunk selection by System~1 with
verification and memory updates in document order by System~2
(Figure~\ref{fig:framework}).

\subsubsection{System 1: Memory-Conditioned Parallel Candidate Selection}
\label{sec:memory-conditioned-proposals}

Here $t$ indexes speculative rounds and $i$ document chunks.
At round~$t$, System~1 independently evaluates a window $\mathcal W_t$
of up to $k$ consecutive chunks in parallel, using the shared query $q$
and memory $m_t$ to select candidate chunks for memory updating:
\begin{equation}
\begin{aligned}
a_{t,i}&\sim\pi_\phi^{\mathrm{sel}}
\left(\cdot\mid q,c_i,m_t\right),
\\
a_{t,i}&\in\{\skipaction{},\update{}\}.
\end{aligned}
\label{eq:memory-proposal-policy}
\end{equation}

The chunks predicted as \update{} form the candidate set
\begin{equation}
\mathcal C_t=\{c_i\in\mathcal W_t\mid a_{t,i}=\update{}\},
\label{eq:candidate-chunk-set}
\end{equation}
which is forwarded to System~2.

\subsubsection{System 2: Speculative Verification and Memory Consolidation}
\label{sec:speculative-verification}

Inspired by speculative decoding's acceptance rule
\citep{leviathan2023speculative,chen2023speculative}, System~2's binary
policy $\pi_\theta^{\mathrm{ver}}$ assesses whether selected chunks warrant
memory updates. For candidate indices $\mathcal I_t=\{i\mid c_i\in\mathcal C_t\}$,
the action-level acceptance probability is
\begin{equation}
\alpha_{t,i}=\min\!\left(1,\,
\frac{\pi_\theta^{\mathrm{ver}}(\update{}\mid q,c_i,m_t)}
{\pi_\phi^{\mathrm{sel}}(\update{}\mid q,c_i,m_t)}\right).
\label{eq:workflow-acceptance}
\end{equation}
Independently draw $\xi_{t,i}\sim\mathcal U(0,1)$ for each candidate
and accept it if $\xi_{t,i}\leq\alpha_{t,i}$.

If any candidate is accepted, commit the first accepted chunk in
document order and generate its replacement memory:
\begin{equation}
m_{t+1}\sim\pi_\theta^{\mathrm{mem}}
(\cdot\mid q,c_{i_t^\star},m_t),
\label{eq:ordered-commitment}
\end{equation}
where $i_t^\star=\min\{i\in\mathcal I_t\mid\xi_{t,i}\leq\alpha_{t,i}\}$.
As a special case, if no candidate is accepted, retain the current memory
and advance past the window. Otherwise, discard remaining
predictions conditioned on $m_t$ and resume immediately after the committed
chunk with $m_{t+1}$. Answering uses the query and final memory;
Algorithm~\ref{alg:memduet-round} in Appendix~\ref{app:inference-algorithm}
gives the complete transitions.

%% file: sections/training_memduet_en.tex
\subsection{Training of MemDuet}
\label{sec:smpi}

\begin{figure}[!t]
\centering
\includegraphics[width=\columnwidth]{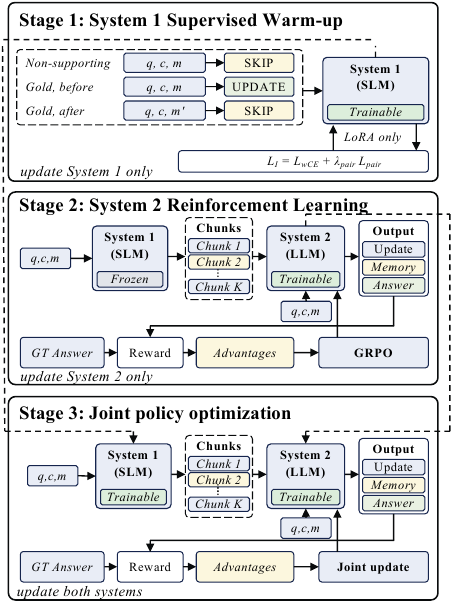}
\caption{Three-stage training of MemDuet. Stage~I trains System~1 with
weighted classification and a paired-memory hinge loss. Stage~II freezes
System~1 and optimizes System~2 with GRPO. Stage~III jointly optimizes
both systems using task rewards.}
\label{fig:three-stage-training}
\end{figure}

Figure~\ref{fig:three-stage-training} summarizes three-stage training.
Stage~I trains memory-conditioned evidence selection.
Stage~II freezes System~1 and trains System~2 to verify candidates,
rewrite memory, and generate answers.
Stage~III jointly optimizes both systems with task rewards,
coordinating selection with verification and memory updates.

\subsubsection{Stage I: System 1 Supervised Warm-up}
\label{sec:training-stage-one}

We initialize System~1 on HotpotQA \citep{yang2018hotpotqa} by aligning
gold supporting facts with document chunks. Evidence-containing chunks
are labeled \update{}, and other chunks \skipaction{}.

For each \update{} example $(q,c,m)$, let $m'$ be the memory after
incorporating the evidence from $c$. We retain the original example
and add a post-update negative:
\begin{equation}
\begin{aligned}
(q,c,m)&\longrightarrow\update{},\\
(q,c,m')&\longrightarrow\skipaction{}.
\end{aligned}
\label{eq:stage-one-data}
\end{equation}
We train on the original examples and post-update negatives using the
memory-contrastive objective:
\begin{equation}
\begin{aligned}
\mathcal L_{\mathrm{Stage\,I}}(\phi)
={}&\mathcal L_{\mathrm{wCE}}+\lambda_{\mathrm{pair}}\mathcal L_{\mathrm{pair}},\\
\mathcal L_{\mathrm{pair}}
={}&\mathbb E_{\mathcal P}\bigl[\max(0,\gamma-\Delta s_\phi)\bigr].
\end{aligned}
\label{eq:stage-one-objective}
\end{equation}
Here, $\mathcal L_{\mathrm{wCE}}$ is class-weighted action cross-entropy;
$\mathcal P$ contains matched memory
pairs, with $\Delta s_\phi=s_\phi(q,c,m)-s_\phi(q,c,m')$.
The score $s_\phi$ is the UPDATE-minus-SKIP logit difference, and
$\gamma$ is the margin. Hyperparameters are reported in
Appendix~\ref{app:training-configuration}.

\subsubsection{Stage II: System 2 Reinforcement Learning}
\label{sec:training-stage-two}

We freeze System~1 at $\phi_{\mathrm{SFT}}$ and optimize all System~2
parameters $\theta$ by reinforcement learning over complete trajectories.
Selections are recomputed online as System~2 updates memory.

For each training instance $(q,D,y^\ast)$, sample $G$ complete trajectories
using the behavior-policy snapshot $\theta_{\mathrm{old}}$.
From the final-answer rewards $R_g=R_{\mathrm{task}}(y_g,y^\ast)$,
we compute the group-relative advantage:
\begin{equation}
\widehat A_g=
\frac{R_g-\operatorname{mean}(R_1,\ldots,R_G)}
{\operatorname{std}(R_1,\ldots,R_G)+\epsilon_{\mathrm{num}}}.
\label{eq:stage-two-advantage}
\end{equation}

Here $g$ indexes trajectories and $\epsilon_{\mathrm{num}}$ is a numerical
stability term.
We use a clipped GRPO loss with generated-token KL regularization
\citep{shao2024deepseekmath}:
\begin{equation}
\begin{aligned}
&\mathcal L_{\mathrm{Stage\,II}}(\theta)
=-\mathbb E_g\Bigl[
\tfrac{1}{|\mathcal E_g|}
\sum\nolimits_{e\in\mathcal E_g}\\
&\qquad\min\bigl(\rho_{g,e}\widehat A_g,
\operatorname{clip}(\rho_{g,e},1-\epsilon,1+\epsilon)
\widehat A_g\bigr)\Bigr]\\
&\quad+\beta\mathbb E_g\Bigl[
\tfrac{1}{|\mathcal T_g|}
\sum\nolimits_{e\in\mathcal T_g}
D_{\mathrm{KL}}\bigl(
\pi_\theta^{\mathrm{gen}}
\,\Vert\,
\pi_{\mathrm{ref}}^{\mathrm{gen}}
\bigr)\Bigr],
\end{aligned}
\label{eq:stage-two-objective}
\end{equation}
where $\mathcal E_g$ contains verification commitment events for nonempty
candidate sets and all generated tokens;
$\mathcal T_g\subseteq\mathcal E_g$ contains only generated tokens.
$e$ is a verification commitment event or generated token, and $\rho_{g,e}$
is its current-to-behavior probability ratio. Verification commitment,
token generation, and Stage~III selection ratios are denoted by
$\rho_{g,t}^{\mathrm{ver}}$, $\rho_{g,t,\ell}^{\mathrm{gen}}$, and
$\rho_{g,t,i}^{\mathrm{sel}}$, respectively; $\epsilon$ is the clipping parameter.
\input{sections/training_event_ratios}

For a round-$t$ verification event with a committed chunk, let
$i_{g,t}^\star$ record its index. The ratio is
\stageTwoVerificationRatio
where $e$ is the round-$t$ verification commitment event.
$\Pr_{\phi,\theta}$ denotes the distribution of commitment outcomes
induced by independent acceptance tests under the given query, window,
candidate indices, and memory. For a candidate index $j\in\mathcal I_t$,
its commitment probability is
$\alpha_{t,j}\prod_{i\in\mathcal I_t,\;i<j}(1-\alpha_{t,i})$.
As a special case, the no-commitment probability is
$\prod_{i\in\mathcal I_t}(1-\alpha_{t,i})$; it is one for empty
$\mathcal I_t$, with no verification-policy loss.
The controller's independent acceptance tests and ordered commitment induce
this distribution; selection and verification calls condition on $(q,c_i,m_t)$.

For a committed-chunk memory update, let $\ell$ index tokens and
$m_{g,t+1,<\ell}$ denote the generated prefix. Its token-level ratio is
\stageTwoMemoryRatio
where $e$ is the generation event for token $m_{g,t+1,\ell}$ and
$\rho_{g,e}=\rho_{g,t,\ell}^{\mathrm{gen}}$.
As a special case, first-chunk initialization tokens $m_{g,0,\ell}$ and
final-answer tokens $y_{g,\ell}$ use the token-level ratio in
Eq.~\eqref{eq:stage-two-explicit-ratios}, conditioned on
$(q,c_1,m_{\mathrm{init}},m_{g,0,<\ell})$ and
$(q,m_{g,\mathrm{fin}},y_{g,<\ell})$, respectively.
Here, $m_{\mathrm{init}}$ is the fixed initial-memory marker and
$m_{g,\mathrm{fin}}$ is the final memory. Both token types contribute to the
policy loss and KL regularizer in Eq.~\eqref{eq:stage-two-objective}.

KL uses full token conditions and $\pi_{\mathrm{ref}}^{\mathrm{gen}}$,
a frozen copy of the initial System~2 model. It applies only to generated
tokens (Appendix~\ref{app:training-stage-two}).

\input{sections/table1_acl_integrated_en}

\subsubsection{Stage III: Joint Policy Optimization}
\label{sec:training-stage-three}

Let $P_{\phi,\theta}(\tau\mid q,D)$ denote the complete trajectory
distribution induced by selection, verification commitments, and
System~2's memory and answer generation
(Appendix~\ref{app:training-stage-three}). With both backbones frozen,
we jointly optimize only their LoRA adapters:
\begin{equation}
\max_{\phi,\theta}\,
\mathbb E_{\tau\sim P_{\phi,\theta}(\cdot\mid q,D)}
\left[R_{\mathrm{task}}(y,y^\ast)\right],
\label{eq:stage-three-objective}
\end{equation}
using fresh rollout groups under $(\phi_{\mathrm{old}},\theta_{\mathrm{old}})$
and the Stage-II advantages. The clipped loss adds selection events
recording both \skipaction{} and \update{}
actions $a_{g,t,i}$, with ratios
\input{sections/training_joint_event_ratios}
where $e$ is the selection event for chunk $i$ in round $t$.
Commitment ratios follow Eq.~\eqref{eq:stage-two-ratio}, replacing
$\phi_{\mathrm{SFT}}$ with $\phi$ in the numerator and
$\phi_{\mathrm{old}}$ in the denominator, retaining dependence on both
systems through Eq.~\eqref{eq:workflow-acceptance}. Generation ratios
and generation-only KL, including the same frozen reference policy,
are retained from Stage~II.

%% file: sections/training_event_ratios.tex
\newcommand{\stageTwoVerificationRatio}{%
\begin{equation}
\begin{aligned}
&\rho_{g,e}=\rho_{g,t}^{\mathrm{ver}}=\\
&\frac{\Pr_{\phi_{\mathrm{SFT}},\theta}
(i_t^\star=i_{g,t}^\star\mid q,\mathcal W_{g,t},\mathcal I_{g,t},m_{g,t})}
{\Pr_{\phi_{\mathrm{SFT}},\theta_{\mathrm{old}}}
(i_t^\star=i_{g,t}^\star\mid q,\mathcal W_{g,t},\mathcal I_{g,t},m_{g,t})},
\end{aligned}
\label{eq:stage-two-ratio}
\end{equation}%
}

\newcommand{\stageTwoMemoryRatio}{%
\begin{equation}
\begin{aligned}
&\rho_{g,e}=\rho_{g,t,\ell}^{\mathrm{gen}}=\\
&\frac{\pi_\theta^{\mathrm{gen}}
(m_{g,t+1,\ell}\mid q,c_{i_{g,t}^\star},m_{g,t},m_{g,t+1,<\ell})}
{\pi_{\theta_{\mathrm{old}}}^{\mathrm{gen}}
(m_{g,t+1,\ell}\mid q,c_{i_{g,t}^\star},m_{g,t},m_{g,t+1,<\ell})},
\end{aligned}
\label{eq:stage-two-explicit-ratios}
\end{equation}%
}

%% file: sections/table1_acl_integrated_en.tex
\begingroup
\definecolor{oursrow}{HTML}{EAF2FB}
\newcommand{\tableoneReproducedAcc}[1]{\accnum{#1}$^{*}$}
\newcommand{\tableoneCaption}{
Accuracy and inference efficiency on RULER-HQA across context lengths
from 7K to 3.5M tokens.
Relative to RL-MemAgent, speedup ($\uparrow$) is
$T_{\mathrm{base}}/T_{\mathrm{method}}$, and the output-token ratio
($\downarrow$) is
$O_{\mathrm{method}}/O_{\mathrm{base}}$, where $T$ and $O$ denote
inference time and generated-token count, respectively.
LycheeMemory ($-G$/$+G$) disables/enables its native memory gate.
$^{*}$ denotes results from our reproduction.
\textbf{Bold} and \underline{underlined} values indicate the best and second-best
reported results at each length within each metric block.
``N/A'' denotes unavailable results; ``-'' denotes results not reported in the original paper.
}
\input{sections/table1_acl_main_horizontal}
\endgroup

%% file: sections/table1_acl_main_horizontal.tex
\providecommand{\tableoneMethod}{Method}
\providecommand{\tableonePerformance}{Accuracy (\%, $\uparrow$)}
\providecommand{\tableoneSpeed}{Time speedup (MemAgent = $1\times$, $\uparrow$)}
\providecommand{\tableoneTokens}{Output tokens (MemAgent = $1\times$, $\downarrow$)}
\providecommand{\tableoneReproducedAcc}[1]{\accnum{#1}}
\begin{table*}[!tp]
\centering
\begingroup
\definecolor{tableoneblue}{HTML}{214E97}
\definecolor{tableoneband}{HTML}{F1F3F7}
\definecolor{tableoneours}{HTML}{EAF1FA}
\definecolor{tableonetext}{HTML}{222222}
\definecolor{tableoneoutline}{HTML}{6C757E}
\color{tableonetext}
\fontsize{8.5}{10}\selectfont
\setlength{\tabcolsep}{0pt}
\renewcommand{\arraystretch}{1.22}
\newcommand{\nodata}{\textemdash}
\newcommand{\accnum}[1]{{\fontsize{8}{9.5}\selectfont #1}}
\newcommand{\modelname}[2]{\mbox{#1\kern0.8pt\raisebox{-0.45ex}{\fontsize{7.5}{8.5}\selectfont #2}}}
\newcommand{\best}[1]{\textbf{#1}}
\newcommand{\runnerup}[1]{\underline{#1}}
\newcommand{\speednum}[1]{{\fontsize{8}{9.5}\selectfont #1$\times$}}
\newcommand{\speedest}[1]{{\fontsize{8}{9.5}\selectfont #1$\times$}}
\newcommand{\ratioval}[1]{{\fontsize{8}{9.5}\selectfont #1$\times$}}
\newcommand{\oursacc}[1]{\textcolor{tableoneblue}{\accnum{#1}}}
\newcommand{\oursspeed}[1]{\textcolor{tableoneblue}{\speednum{#1}}}
\newcommand{\oursratio}[1]{\textcolor{tableoneblue}{\ratioval{#1}}}
\input{sections/table1_total_output_token_cells}
\begin{tabularx}{\textwidth}{@{}l*{10}{>{\centering\arraybackslash}X}@{}}
\arrayrulecolor{tableoneoutline}\toprule
\textbf{\tableoneMethod} & \textbf{7K} & \textbf{14K} & \textbf{28K} & \textbf{56K} & \textbf{112K} & \textbf{224K} & \textbf{448K} & \textbf{896K} & \textbf{1.75M} & \textbf{3.5M} \\
\arrayrulecolor{black}\midrule
\rowcolor{tableoneband}\multicolumn{11}{c}{\textcolor{tableoneblue}{\itshape\tableonePerformance}} \\
\midrule
\modelname{Search-R1}{7B} & \accnum{72.66} & \accnum{71.88} & \accnum{67.71} & \accnum{73.96} & \accnum{66.67} & \accnum{62.50} & \accnum{64.58} & \accnum{67.71} & \accnum{67.19} & \nodata \\
\midrule
\modelname{QwenLong-L1}{32B} & \accnum{72.66} & \accnum{75.00} & \accnum{72.66} & \accnum{60.94} & \accnum{31.25} & \accnum{17.19} & \accnum{13.28} & \accnum{11.72} & \accnum{N/A} & \accnum{N/A} \\
\modelname{DS-Distill}{32B} & \accnum{70.31} & \accnum{66.41} & \accnum{65.62} & \accnum{46.88} & \accnum{23.44} & \accnum{13.28} & \accnum{7.81} & \accnum{7.03} & \accnum{N/A} & \accnum{N/A} \\
\midrule
\modelname{RL-MemAgent}{7B} & \accnum{81.25} & \accnum{81.25} & \accnum{\runnerup{82.03}} & \accnum{\runnerup{80.47}} & \accnum{79.69} & \accnum{75.78} & \accnum{76.56} & \accnum{74.22} & \accnum{77.34} & \accnum{71.88} \\
\modelname{LycheeMemory ($-G$)}{7B} & \accnum{80.47} & \accnum{81.25} & \accnum{\runnerup{82.03}} & \accnum{\best{81.25}} & \accnum{80.47} & \accnum{\runnerup{79.69}} & \accnum{75.00} & \accnum{75.78} & \accnum{\runnerup{78.12}} & \nodata \\
\modelname{LycheeMemory ($+G$)}{7B} & \accnum{77.34} & \accnum{76.56} & \accnum{75.00} & \accnum{76.56} & \accnum{75.78} & \accnum{73.44} & \accnum{74.22} & \accnum{72.66} & \accnum{71.09} & \nodata \\
\modelname{ReMemR1}{7B} & \tableoneReproducedAcc{\runnerup{82.03}} & \tableoneReproducedAcc{\runnerup{82.81}} & \tableoneReproducedAcc{81.25} & \tableoneReproducedAcc{78.91} & \tableoneReproducedAcc{\runnerup{82.81}} & \tableoneReproducedAcc{\runnerup{79.69}} & \tableoneReproducedAcc{\runnerup{80.47}} & \tableoneReproducedAcc{\runnerup{77.34}} & \tableoneReproducedAcc{71.09} & \tableoneReproducedAcc{\runnerup{72.66}} \\
\midrule
\rowcolor{tableoneours}\textcolor{tableoneblue}{\modelname{\textbf{MemDuet}}{7B}} & \oursacc{\best{87.50}} & \oursacc{\best{83.59}} & \oursacc{\best{89.06}} & \oursacc{79.69} & \oursacc{\best{87.50}} & \oursacc{\best{85.94}} & \oursacc{\best{82.81}} & \oursacc{\best{82.03}} & \oursacc{\best{79.69}} & \oursacc{\best{75.78}} \\
\midrule
\rowcolor{tableoneband}\multicolumn{11}{c}{\textcolor{tableoneblue}{\itshape\tableoneSpeed}} \\
\midrule
\modelname{LycheeMemory ($-G$)}{7B} & \speedest{\runnerup{0.191}} & \speedest{0.206} & \speednum{0.233} & \speedest{0.305} & \speednum{0.377} & \speedest{0.445} & \speedest{0.393} & \speedest{0.515} & \speedest{0.411} & \speedest{0.399} \\
\modelname{LycheeMemory ($+G$)}{7B} & \speedest{0.19} & \speedest{\runnerup{0.25}} & \speedest{\runnerup{0.37}} & \speedest{\runnerup{0.76}} & \speedest{\runnerup{1.32}} & \speedest{\runnerup{2.62}} & \speedest{\runnerup{3.81}} & \speedest{\runnerup{9.12}} & \speedest{\runnerup{11.6}} & \nodata \\
\modelname{ReMemR1}{7B} & \speedest{0.173} & \speedest{0.185} & \speedest{0.213} & \speedest{0.272} & \speedest{0.325} & \speedest{0.353} & \speedest{0.413} & \speedest{0.469} & \speedest{0.456} & \speedest{\runnerup{0.560}} \\
\midrule
\rowcolor{tableoneours}\textcolor{tableoneblue}{\modelname{\textbf{MemDuet}}{7B}} & \oursspeed{\best{0.983}} & \oursspeed{\best{1.445}} & \oursspeed{\best{1.588}} & \oursspeed{\best{3.541}} & \oursspeed{\best{5.885}} & \oursspeed{\best{9.885}} & \oursspeed{\best{11.913}} & \oursspeed{\best{18.688}} & \oursspeed{\best{39.085}} & \oursspeed{\best{49.613}} \\
\midrule
\rowcolor{tableoneband}\multicolumn{11}{c}{\textcolor{tableoneblue}{\itshape\tableoneTokens}} \\
\midrule
\modelname{LycheeMemory ($-G$)}{7B} & \ratioval{\runnerup{1.99}} & \ratioval{\runnerup{1.86}} & \ratioval{\runnerup{1.54}} & \ratioval{\runnerup{1.21}} & \ratioval{\runnerup{0.92}} & \ratioval{\runnerup{0.81}} & \ratioval{\runnerup{0.92}} & \ratioval{\runnerup{0.70}} & \ratioval{\runnerup{0.89}} & \ratioval{\runnerup{0.90}} \\
\modelname{ReMemR1}{7B} & \ratioval{4.75} & \ratioval{4.40} & \ratioval{3.62} & \ratioval{2.82} & \ratioval{2.29} & \ratioval{2.11} & \ratioval{1.78} & \ratioval{1.57} & \ratioval{1.62} & \ratioval{1.28} \\
\midrule
\rowcolor{tableoneours}\textcolor{tableoneblue}{\modelname{\textbf{MemDuet}}{7B}} & \tableoneTotalOutputCells \\
\arrayrulecolor{tableoneoutline}\bottomrule
\end{tabularx}
\endgroup
\caption{\tableoneCaption}
\label{tab:main_results}
\end{table*}

%% file: sections/table1_total_output_token_cells.tex
\newcommand{\tableoneTotalOutputCells}{%
  \oursratio{\best{0.9050}} & \oursratio{\best{0.5659}} & \oursratio{\best{0.3671}} & \oursratio{\best{0.2012}} & \oursratio{\best{0.1154}} & \oursratio{\best{0.0607}} & \oursratio{\best{0.0543}} & \oursratio{\best{0.0460}} & \oursratio{\best{0.0117}} & \oursratio{\best{0.0086}}%
}

%% file: sections/training_joint_event_ratios.tex
\begin{equation}
\begin{aligned}
\rho_{g,e}=\rho_{g,t,i}^{\mathrm{sel}}
&=\frac{\pi_\phi^{\mathrm{sel}}(a_{g,t,i}\mid q,c_i,m_{g,t})}
{\pi_{\phi_{\mathrm{old}}}^{\mathrm{sel}}(a_{g,t,i}\mid q,c_i,m_{g,t})},
\end{aligned}
\label{eq:stage-three-event-ratios}
\end{equation}

%% file: sections/integrated_experiments_en.tex
\input{sections/experimental_setup_en}

\subsection{Main Results}

\paragraph{Length scaling on RULER-HQA.}
Table~\ref{tab:main_results} shows that MemDuet achieves an increasingly
favorable accuracy--efficiency trade-off as context grows.
At 3.5M tokens, it outperforms RL-MemAgent by 3.90 points while achieving
a 49.61$\times$ speedup and generating 99.14\% fewer output tokens.
At 7K, the limited number of chunks provides little opportunity to offset
the fixed cost of System~1 scanning and cross-system coordination.
As the context grows, this overhead is amortized and more unnecessary
System~2 processing can be avoided, making the efficiency gains
increasingly pronounced without sacrificing accuracy.

\paragraph{Generalization to realistic long-context tasks.}
Table~\ref{tab:additional_results} evaluates MemDuet on LongBench v2,
which contains realistic long-context tasks, and on length-controlled
2WikiMultihopQA. MemDuet achieves the highest option accuracy on the Long and Ultra groups of
LongBench v2, with 3.00$\times$ and 8.03$\times$ speedups.
Its lower option accuracy on the Short and Medium groups is consistent with selective
processing occasionally discarding useful evidence when the full input
is still manageable.
The same transition appears in sub-EM on 2WikiMultihopQA: MemDuet trails at 14K
and 56K, but leads at 224K and 896K, indicating that its advantages are
concentrated in the long-context regime for which it is designed.

\subsection{Analysis}
\label{sec:experimental-analysis}
\input{sections/analysis_integrated_en}

%% file: sections/experimental_setup_en.tex
\subsection{Experimental Setup}
\label{sec:experimental-setup}

\paragraph{Datasets and evaluation.}
We follow MemAgent's RULER-HQA construction (7K--3.5M tokens)
\citep{yu2026memagent}, reserving its 128-question development set for
testing and a disjoint 128-question HotpotQA validation set for checkpoint
selection.
We further evaluate on LongBench v2
\citep{bai2025longbenchv2} and a long-context variant of
2WikiMultihopQA \citep{ho2020twowiki}.
We report option accuracy on LongBench v2 and normalized substring
exact match (sub-EM) on RULER-HQA and 2WikiMultihopQA.
To assess inference efficiency, we additionally report speedup
relative to MemAgent and the number of generated output tokens.
Dataset construction, sample sizes, and length groupings are detailed
in Appendix~\ref{app:data-preprocessing}; evaluation protocols are
provided in Appendix~\ref{app:metrics-protocol}.

\paragraph{Baselines.}
We compare MemDuet with Search-R1-7B
\citep{jin2025searchr1}, QwenLong-L1-32B
\citep{wan2025qwenlong}, DS-Distill-32B
\citep{deepseek2025r1}, RL-MemAgent-7B
\citep{yu2026memagent}, LycheeMemory-7B
\citep{chen2026lycheememory}, and ReMemR1-7B
\citep{shi2025rememr1}.

\paragraph{Implementation.}
We use Qwen3.5-2B as System~1 for the main results and evaluate
Qwen2.5-0.5B, Qwen2.5-1.5B, and Qwen3.5-0.8B in the System~1
backbone comparison \citep{qwen2025qwen25,qwen2026qwen35}.
System~2 uses the same Qwen2.5-7B-Instruct backbone as MemAgent
\citep{qwen2025qwen25,yu2026memagent}.
See Appendix~\ref{app:training-implementation} for training and implementation.

%% file: sections/analysis_integrated_en.tex
\begingroup
\newcommand{\analysislang}[1]{#1}
\input{sections/analysis_bundle_tables}
\input{sections/analysis_bundle_text_en}
\endgroup

%% file: sections/analysis_bundle_tables.tex
\newcommand{\lang}[1]{#1}
\newcommand{\analysislabel}[1]{\label{#1}}
\newcommand{\analysisref}[1]{\ref{#1}}
\definecolor{memblue}{HTML}{2A5AA5}
\definecolor{memshade}{HTML}{EAF1FB}
\colorlet{ablationshade}{memshade}
\newcommand{\blue}[1]{\textcolor{memblue}{#1}}
\newcommand{\tablefont}{\fontsize{7.9}{9.2}\selectfont
  \setlength{\tabcolsep}{2.2pt}\renewcommand{\arraystretch}{1.06}}

\newcommand{\scalingfigure}{%
  \analysislang{\includegraphics[width=\linewidth]{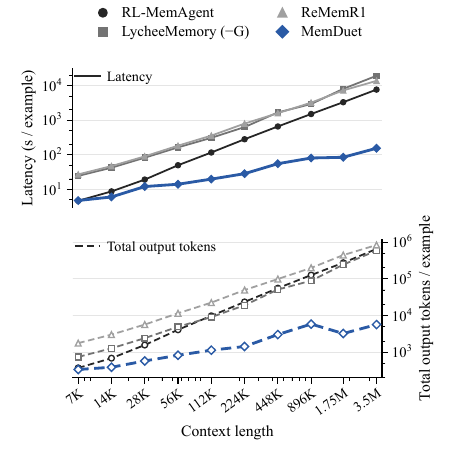}}
  \caption{\lang{Efficiency scaling on RULER-HQA from 7K to 3.5M tokens. The upper and lower panels report latency and total output tokens from both systems per example, respectively, on logarithmic axes.}}
  \analysislabel{fig:efficiency_scaling}}

\newcommand{\modelchoicetable}{%
  {\fontsize{8.8}{10.3}\selectfont
  \setlength{\tabcolsep}{1pt}\renewcommand{\arraystretch}{1.10}
  \begin{tabularx}{\linewidth}{@{}l*{4}{>{\centering\arraybackslash}X}@{}}
  \toprule
  System~1
    & F1 (\%) $\uparrow$
    & Acc. (\%) $\uparrow$ & Speed $\uparrow$
    & Tokens $\downarrow$\\
  \midrule
  \csname @@input\endcsname sections/s1_backbone_summary_rows.tex
  \bottomrule
  \end{tabularx}}
  \caption{\lang{System~1 on RULER-HQA (7K--3.5M tokens). Acc., Speed, and Tokens are ten-length means; the latter two use per-length ratios to RL-MemAgent. F1 measures \update{} selection. Appendix~\ref{app:s1-backbone-results} gives full results and timing normalization.}}
  \analysislabel{tab:s1_choice}}

\newcommand{\concurrencytable}{%
  {\fontsize{7.9}{9.2}\selectfont
  \setlength{\tabcolsep}{1.5pt}\renewcommand{\arraystretch}{1.16}
  \begin{tabularx}{\linewidth}{@{}>{\centering\arraybackslash}p{17mm}
    *{4}{>{\centering\arraybackslash}X}@{}}
  \toprule
  \multirow[c]{2}{*}{$k$}
    & \multicolumn{4}{c}{\textbf{Speed}\,$\uparrow$}\\
  \cmidrule(lr){2-5}
  & \textbf{112K} & \textbf{448K} & \textbf{896K} & \textbf{1.75M}\\
  \midrule
  \concurrencyrow{white}{1}{$1\times$}{$1\times$}{$1\times$}{$1\times$}
  \concurrencyrow{white}{2}{\concurrencybest{1.006}}{\concurrencybest{1.032}}{$1.010\times$}{$1.087\times$}
  \concurrencyrow{memshade}{\blue{4}}{$0.968\times$}{$0.981\times$}{\concurrencybest{1.080}}{$1.107\times$}
  \concurrencyrow{white}{8}{$0.920\times$}{$0.867\times$}{$0.919\times$}{\concurrencybest{1.111}}
  \concurrencyrow{white}{16}{$0.768\times$}{$0.777\times$}{$0.838\times$}{$0.927\times$}
  \concurrencyrow{white}{32}{$0.741\times$}{$0.567\times$}{$0.828\times$}{$0.763\times$}
  \bottomrule
  \end{tabularx}}
  \caption{\lang{Effect of the parallel chunk window width $k$. Speedups are measured relative to the same-length serial configuration ($k=1$); values below $1\times$ indicate slower execution. \textbf{Bold} marks the highest speedup at each length, and shading denotes the default $k=4$.}}
  \analysislabel{tab:parallel_chunks}}

\newcommand{\concurrencybest}[1]{\textbf{#1}$\boldsymbol{\times}$}
\newcommand{\concurrencyrow}[6]{%
  \rowcolor{#1}[1.5pt][1.5pt]#2 & #3 & #4 & #5 & #6\\}

\newcommand{\configurationrow}[6]{%
  \rowcolor{#1}#2 & #3 & #4 & #5 & #6\\}

\newcommand{\ablationtable}{%
  {\fontsize{7.7}{9}\selectfont
  \setlength{\tabcolsep}{2.1pt}\renewcommand{\arraystretch}{1.10}
  \begin{tabularx}{\linewidth}{@{}>{\raggedright\arraybackslash}X|ccc|cc@{}}
  \toprule
  \multicolumn{1}{c|}{\multirow[c]{2}{*}{\textbf{\lang{Framework}}}}
    & \multicolumn{3}{c|}{\textbf{Acc. (\%)} $\uparrow$}
    & \multirow[c]{2}{*}{\textbf{Speed}\,$\uparrow$}
    & \multirow[c]{2}{*}{\textbf{Tokens}\,$\downarrow$}\\
  \cmidrule(lr){2-4}
  & \textbf{112K} & \textbf{448K} & \textbf{896K} & &\\
  \midrule
  \textbf{MemDuet}
    & \textbf{87.50} & \textbf{82.81} & \textbf{82.03}
    & \textbf{\boldmath$12.16\times$} & \textbf{\boldmath$.0719\times$}\\
  \midrule
  \addlinespace[3pt]
  \rowcolor{ablationshade}
  \multicolumn{6}{@{}l@{}}{\hspace{2pt}\rule[-2pt]{0pt}{13pt}\blue{\textbf{\textit{\lang{Inference Mechanisms}}}}}\\
  \midrule
  \hspace{10pt}\lang{w/o parallel execution}
    & 87.50 & 82.81 & 82.03 & $11.84\times$ & $.0661\times$\\
  \hspace{10pt}\lang{w/o draft--verify}
    & 84.38 & 78.91 & 78.13 & $9.89\times$ & $.0769\times$\\
  \hspace{10pt}\lang{w/o state reevaluation}
    & 84.38 & 80.47 & 78.13 & $11.58\times$ & $.0712\times$\\
  \midrule
  \addlinespace[3pt]
  \rowcolor{ablationshade}
  \multicolumn{6}{@{}l@{}}{\hspace{2pt}\rule[-2pt]{0pt}{13pt}\blue{\textbf{\textit{\lang{Three-Stage Training}}}}}\\
  \midrule
  \hspace{10pt}\lang{w/o Stage~I}
    & 82.03 & 78.91 & 77.34 & $2.54\times$ & $.2107\times$\\
  \hspace{10pt}\lang{w/o Stage~II}
    & 38.28 & 25.78 & 29.69 & $28.54\times$ & $.0082\times$\\
  \hspace{10pt}\lang{w/o Stage~III}
    & 81.25 & 80.47 & 78.91 & $8.81\times$ & $.05682\times$\\
  \bottomrule
  \end{tabularx}}
  \caption{\lang{Ablations of MemDuet's inference mechanisms and three-stage training on RULER-HQA. Accuracy is reported at 112K, 448K, and 896K tokens. Speed and Tokens equally average per-length ratios relative to RL-MemAgent. \emph{W/o Stage~I} retains Stages~II and~III; \emph{w/o Stage~II} retains Stages~I and~III; \emph{w/o Stage~III} retains Stages~I and~II.}}
  \analysislabel{tab:ablation}}

\newcommand{\apitable}{%
  {\tablefont
  \begin{tabularx}{\linewidth}{@{}>{\raggedright\arraybackslash}p{17mm}>{\raggedright\arraybackslash}Xccc@{}}
  \toprule
  \lang{API family} & \lang{Method} & \shortstack{\lang{Option Acc.}\\(\%)\,$\uparrow$}
    & \mbox{\lang{Latency} (s)\,$\downarrow$}
    & \mbox{\lang{Cost} (\$)\,$\downarrow$}\\
  \midrule
  DeepSeek-V4 & MemAgent & 57.46 & 634.23 & 0.1974\\
  \rowcolor{memshade}
  & \blue{MemDuet} & 57.65 & 178.55 & 0.0825\\
  \midrule
  GPT-5.6 & MemAgent & 61.63 & 438.92 & 0.5978\\
  \rowcolor{memshade}
  & \blue{MemDuet} & 61.03 & 128.46 & 0.2398\\
  \bottomrule
  \end{tabularx}}
  \caption{\lang{Training-free transfer to proprietary models on LongBench v2, reporting option accuracy. MemDuet combines off-the-shelf System~1 and System~2 APIs using binary accept-or-reject verification. Latency and cost include both systems.}}
  \analysislabel{tab:api_transfer}}

%% file: sections/analysis_bundle_text_en.tex
\begin{figure}[!t]\centering\scalingfigure\end{figure}
\input{sections/table2_integrated_en}

\paragraph{\lang{Efficiency scaling.}}
\lang{Figure~\analysisref{fig:efficiency_scaling} shows that MemDuet's latency
and output tokens grow more slowly with context length than those
of RL-MemAgent. Although MemDuet incurs a small latency overhead at 7K,
its runtime advantage becomes increasingly pronounced on longer inputs.
The shallower growth in output tokens is consistent with restricting
large-model memory updates to selected evidence rather than rewriting
memory for every chunk. Latency savings are smaller than token
savings, since input processing and System~1 computation remain necessary
even when output tokens are substantially reduced.}

\begin{table}[!t]\centering\ablationtable\end{table}

\paragraph{\lang{Inference mechanisms.}}
\lang{\emph{W/o draft--verify} retains System~1 selection but removes
System~2 verification and acceptance tests: selected \update{} actions
directly trigger memory rewriting, followed by reevaluation of later chunks.
\emph{W/o state reevaluation} reuses selection and verification/acceptance
results computed with the window-start memory even after updates,
processing chunks in document order; memory generation still uses the latest memory.
\emph{W/o parallel execution} changes $k=4$ to $k=1$, retaining all
other decision rules.
The $k=1$ variant preserves accuracy, with speedup changing from
$12.16\times$ to $11.84\times$.
Removing draft--verify or state reevaluation reduces accuracy by up to
3.90 points; draft--verify removal also lowers speedup to $9.89\times$.
Verification and state reevaluation thus preserve quality while reducing
large-model computation.}

\paragraph{\lang{Three-stage training.}}
\lang{The three training stages play complementary roles. Without Stage~I,
System~1 tends to propose updates for nearly all chunks, maintaining
coverage of supporting evidence and limiting the accuracy drop to
3.90--5.47 points relative to the full configuration. More frequent
verification and memory rewriting increase computation, reducing speedup
from $12.16\times$ to $2.54\times$ and roughly tripling the mean output-token ratio.
Without Stage~II, System~2 lacks dedicated reinforcement learning for
memory rewriting, substantially weakening evidence consolidation.
The retained Stage~III joint training partially restores this capability,
keeping accuracy above the Qwen2.5-7B-Instruct baseline
(112K: 32.81\%; 448K: 6.25\%; 896K: 1.56\%).
Stage~III aligns the two policies: joint optimization recovers 2.34--6.25
accuracy points over \emph{w/o Stage~III}.
The curriculum thus progressively learns when to update, how to
update, and how to coordinate the two decisions.}

\begin{table}[!tbp]\centering\modelchoicetable\end{table}

\paragraph{\lang{Choice of System~1.}}
\lang{System~1 should favor evidence coverage, since missed evidence cannot be
recovered, whereas false positives can still be filtered by System~2.
Across both Qwen2.5 and Qwen3.5 backbones, MemDuet consistently turns small
models into effective proposers, achieving 80.08--83.36\% downstream
accuracy with mean speedups of $5.69$--$14.26\times$.
Qwen3.5-0.8B has the highest selection F1 and lowest output-token ratio,
while Qwen3.5-2B, used for the main results, attains the highest final
accuracy and mean speedup. Selection quality should be evaluated
alongside answer quality and inference cost.}

\begin{table}[!tbp]\centering\concurrencytable\end{table}

\paragraph{\lang{Effect of parallelism.}}
\lang{Parallel chunk processing provides only modest gains because MemDuet
must preserve the document order of memory updates. System~1 evaluates
chunks concurrently, but System~2 rewrites memory from the first accepted
chunk in document order and reevaluates later chunks under the updated memory.
As $k$ increases, more speculative computation may therefore be discarded,
and synchronization overhead can outweigh the benefit of concurrency.
The best speedup is $1.111\times$ at $k=8$ and 1.75M tokens, whereas $k=16$
and $k=32$ slow inference at every evaluated length. Moderate parallelism
thus provides the best balance between concurrent evidence discovery and
memory consolidation in document order.}

\begin{table}[!tbp]\centering\apitable\end{table}

\paragraph{\lang{Transfer to proprietary models.}}
\lang{MemDuet can be directly applied to proprietary APIs without additional
training: a smaller model serves as System~1 to propose candidate chunks,
while a stronger model serves as System~2 to accept or reject each proposal
before memory rewriting. This plug-and-play setup reduces latency by
70.7--71.8\% and cost by 58.2--59.9\%, with option accuracy changing by at most
0.60 points.}

%% file: sections/table2_integrated_en.tex
\begin{table}[!tbp]
\centering
\begingroup
\newcommand{\tabletwometricheader}{Metric}
\newcommand{\tabletwomethod}{Method}
\newcommand{\tabletwoaccuracy}{\shortstack[l]{Option\\Acc.}}
\newcommand{\tabletwosubem}{sub-EM}
\newcommand{\tabletwospeedlabel}{Speedup}
\newcommand{\tabletwooutput}{Tokens}
\input{sections/table2_integrated_body}
\endgroup
\caption{
LongBench v2 (top; option accuracy, \%) and long-context
2WikiMultihopQA (bottom; sub-EM, \%) results.
LycheeMemory uses no gate ($-G$); \textbf{bold} marks the best values per length and metric.
}
\label{tab:additional_results}
\end{table}

%% file: sections/table2_integrated_body.tex
\begingroup
\definecolor{tabletwoours}{HTML}{EAF1FA}
\definecolor{tabletwometricshade}{HTML}{F1F3F7}
\definecolor{tabletwometricblue}{HTML}{214E97}
\newcommand{\tabletwospeed}[1]{$#1\times$}
\newcommand{\tabletwotokenratio}[1]{$#1\times$}
\newcommand{\tabletwobest}[1]{{\bfseries\boldmath #1}}
\newcommand{\tabletworunnerup}[1]{#1}
\newcommand{\tabletwometrictext}[1]{\textcolor{tabletwometricblue}{\itshape #1}}
\newcommand{\tabletwometriccell}[1]{\cellcolor{tabletwometricshade}\tabletwometrictext{#1}}
\newcommand{\tabletwometricspan}[2]{\cellcolor{tabletwometricshade}\multirow[c]{-#1}{*}{\tabletwometrictext{\mbox{#2}}}}
\newcommand{\tabletwooursname}{\textcolor{tabletwometricblue}{\bfseries MemDuet}}
\newcommand{\tabletwooursbest}[1]{\textcolor{tabletwometricblue}{\tabletwobest{#1}}}
\newcommand{\tabletwooursplain}[1]{\textcolor{tabletwometricblue}{#1}}
\fontsize{8}{9.5}\selectfont
\setlength{\tabcolsep}{0.7pt}
\renewcommand{\arraystretch}{1.17}
\begin{tabularx}{\columnwidth}{@{}p{34pt}l*{4}{>{\centering\arraybackslash}X}@{}}
\toprule
\multicolumn{6}{c}{\textbf{LongBench v2}} \\
\tabletwometricheader & \tabletwomethod & Short & Medium & Long & Ultra \\
\midrule
\tabletwometriccell{} & MemAgent & \tabletwobest{33.62} & \tabletwobest{31.91} & 25.00 & 28.28 \\
\tabletwometriccell{} & LycheeMemory ($-G$) & \tabletworunnerup{25.86} & \tabletwobest{31.91} & 29.00 & 28.28 \\
\rowcolor{tabletwoours}
\tabletwometricspan{3}{\tabletwoaccuracy\,$\uparrow$} & \tabletwooursname & \tabletwooursplain{31.03} & \tabletwooursplain{31.38} & \tabletwooursbest{30.00} & \tabletwooursbest{29.29} \\
\midrule
\tabletwometriccell{} & LycheeMemory ($-G$) & \tabletworunnerup{\tabletwospeed{0.45}} & \tabletworunnerup{\tabletwospeed{0.54}} & \tabletworunnerup{\tabletwospeed{0.55}} & \tabletworunnerup{\tabletwospeed{0.59}} \\
\rowcolor{tabletwoours}
\tabletwometricspan{2}{\tabletwospeedlabel\,$\uparrow$} & \tabletwooursname & \tabletwooursbest{\tabletwospeed{1.69}} & \tabletwooursbest{\tabletwospeed{1.96}} & \tabletwooursbest{\tabletwospeed{3.00}} & \tabletwooursbest{\tabletwospeed{8.03}} \\
\midrule
\tabletwometriccell{} & LycheeMemory ($-G$) & \tabletworunnerup{\tabletwotokenratio{1.10}} & \tabletworunnerup{\tabletwotokenratio{0.90}} & \tabletworunnerup{\tabletwotokenratio{0.89}} & \tabletworunnerup{\tabletwotokenratio{0.83}} \\
\rowcolor{tabletwoours}
\tabletwometricspan{2}{\tabletwooutput\,$\downarrow$} & \tabletwooursname & \tabletwooursbest{\tabletwotokenratio{0.52}} & \tabletwooursbest{\tabletwotokenratio{0.42}} & \tabletwooursbest{\tabletwotokenratio{0.27}} & \tabletwooursbest{\tabletwotokenratio{0.09}} \\
\midrule
\multicolumn{6}{c}{\textbf{2WikiMultiHopQA}} \\
\tabletwometricheader & \tabletwomethod & 14K & 56K & 224K & 896K \\
\midrule
\tabletwometriccell{} & MemAgent & \tabletwobest{67.97} & \tabletwobest{67.19} & \tabletworunnerup{57.81} & 57.03 \\
\tabletwometriccell{} & LycheeMemory ($-G$) & \tabletworunnerup{64.84} & \tabletworunnerup{64.84} & 59.38 & 52.34 \\
\rowcolor{tabletwoours}
\tabletwometricspan{3}{\tabletwosubem\,$\uparrow$} & \tabletwooursname & \tabletwooursplain{64.84} & \tabletwooursplain{61.72} & \tabletwooursbest{61.72} & \tabletwooursbest{59.38} \\
\midrule
\tabletwometriccell{} & LycheeMemory ($-G$) & \tabletworunnerup{\tabletwospeed{0.27}} & \tabletworunnerup{\tabletwospeed{0.42}} & \tabletworunnerup{\tabletwospeed{0.63}} & \tabletworunnerup{\tabletwospeed{0.64}} \\
\rowcolor{tabletwoours}
\tabletwometricspan{2}{\tabletwospeedlabel\,$\uparrow$} & \tabletwooursname & \tabletwooursbest{\tabletwospeed{1.46}} & \tabletwooursbest{\tabletwospeed{3.56}} & \tabletwooursbest{\tabletwospeed{5.56}} & \tabletwooursbest{\tabletwospeed{7.35}} \\
\midrule
\tabletwometriccell{} & LycheeMemory ($-G$) & \tabletworunnerup{\tabletwotokenratio{1.93}} & \tabletworunnerup{\tabletwotokenratio{1.20}} & \tabletworunnerup{\tabletwotokenratio{0.77}} & \tabletworunnerup{\tabletwotokenratio{0.77}} \\
\rowcolor{tabletwoours}
\tabletwometricspan{2}{\tabletwooutput\,$\downarrow$} & \tabletwooursname & \tabletwooursbest{\tabletwotokenratio{0.60}} & \tabletwooursbest{\tabletwotokenratio{0.21}} & \tabletwooursbest{\tabletwotokenratio{0.14}} & \tabletwooursbest{\tabletwotokenratio{0.11}} \\
\bottomrule
\end{tabularx}
\endgroup

%% file: sections/appendices_en.tex
\newcommand{\apptext}[1]{#1}
\raggedbottom

\section{Experimental Setup}
\label{app:experimental-setup}
\label{app:evaluation-details}

\subsection{Dataset Construction}
\label{app:data-preprocessing}

\begin{itemize}[leftmargin=*,topsep=3pt,itemsep=4pt,parsep=0pt]
\item \textbf{RULER-HQA}: We combine the RULER evaluation framework \citep{hsieh2024ruler},
HotpotQA validation questions \citep{yang2018hotpotqa}, and MemAgent's
controlled-length document construction \citep{yu2026memagent}.
Supporting documents are combined with distractors. The test sets retain
the same 128 questions and supporting evidence across length settings,
while additional documents increase the input length. Testing covers
ten nominal lengths: 7K, 14K, 28K, 56K, 112K, 224K, 448K,
896K, 1.75M, and 3.5M tokens. Documents are divided into consecutive
5,000-token chunks using the updater tokenizer, preserving their order.
We reserve the 128 questions in MemAgent's released development set
exclusively for final testing. A separate set of 128 questions from the
remaining HotpotQA validation split is used for training-time validation
and checkpoint selection (Appendix~\ref{app:training-stage-two}).

\item \textbf{LongBench v2}: Table~\ref{tab:additional_results} uses our four
token-length groups, covering all 503 questions. Let $L$ be the
source-context token count measured with the fixed Qwen2.5-7B-Instruct
tokenizer:
Short ($L<32{,}768$; 116 questions),
Medium ($32{,}768\le L<131{,}072$; 188),
Long ($131{,}072\le L\le262{,}144$; 100), and
Ultra ($L>262{,}144$; 99).

\item \textbf{2WikiMultihopQA}: We use the same evidence-plus-distractor
construction strategy as RULER-HQA, following MemAgent \citep{yu2026memagent}.
The data come from FlashRAG \citep{jin2025flashrag}.
We exclude questions missing annotated
supporting documents and build a deduplicated title-and-text document pool.
Each question retains its reference answers and original context, including
all supporting documents. Additional documents are sampled from the pool,
excluding that question's original documents, to reach 100, 400, 1,600, or
6,400 documents. Each context is shuffled and concatenated with
numbered document headers. These counts define the nominal 14K, 56K, 224K,
and 896K settings in Table~\ref{tab:additional_results}.

All lengths and methods share the same 128 questions: the first 128 of
640 sampled eligible question indices.
\end{itemize}

\subsection{Models and Baselines}
\label{app:table-provenance}

\begin{itemize}[leftmargin=*,topsep=3pt,itemsep=4pt,parsep=0pt]
\item \textbf{System~1}: We use Qwen2.5-0.5B, Qwen2.5-1.5B,
Qwen3.5-0.8B, and Qwen3.5-2B as small-model backbones
\citep{qwen2025qwen25,qwen2026qwen35}.
\item \textbf{System~2}: We use Qwen2.5-7B-Instruct
as the large-model backbone
\citep{qwen2025qwen25}.
\end{itemize}

We group the baselines in the complete comparison
(Table~\ref{tab:discussion-length-scaling}) by their long-context
processing mechanism:
\begin{itemize}[leftmargin=*,topsep=3pt,itemsep=4pt,parsep=0pt]
\item \textbf{Base-model approaches}: Qwen2.5-Instruct (7B, 14B, and 32B)
\citep{qwen2025qwen25}, Qwen2.5-1M (7B and 14B)
\citep{yang2025qwen25_1m},
QwenLong-L1 (32B), and DS-Distill (7B, 14B, and 32B) generate answers
directly from the supplied context, without an external retrieval or
memory-update workflow.
\item \textbf{RAG-based approaches}: RAG and Search-R1, both using
Qwen2.5-7B-Instruct, retrieve relevant documents before or during
answer generation.
\item \textbf{Agent-based approaches}: RL-MemAgent (7B and 14B),
InfMem (4B), GRU-Mem (7B), LycheeMemory (7B, $-G$ and $+G$),
ReMemR1 (7B), and
QwenLong-L1.5-30B-A3B \citep{shen2025qwenlongl15}
maintain and update memory to process long contexts incrementally.
\end{itemize}

\subsection{Evaluation Metrics and Protocol}
\label{app:metrics-protocol}

\paragraph{Open-ended question answering.}
For RULER-HQA and long-context 2WikiMultihopQA, we report normalized
substring exact match (sub-EM). We lowercase predictions and reference
answers, remove punctuation and English articles, and collapse whitespace.
Following the MemAgent scorer, a nonempty extracted prediction receives
1 if either normalized answer is a substring of the other, and 0 otherwise.
We take the maximum score over reference-answer aliases and report the
mean as a percentage.

\paragraph{Multiple-choice question answering.}
For LongBench v2, we report option accuracy: the percentage of questions
whose extracted prediction in \{A, B, C, D\} exactly matches the reference
option. Missing or unparseable option labels score 0.
This compares option labels without the open-ended QA normalization or
substring-matching rule. The question-and-options prompt and final-answer
format are given in Appendix~\ref{app:prompt-longbench-v2}.

\paragraph{Inference time and speedup.}
Local per-question latency covers initialization, selection, verification,
committed rewrites, and answering, excluding model loading, warm-up, and
offline input preparation. Rejected or stale requests may still incur
processing cost. Within one length setting and the same question cohort,
let $T_{\mathrm{MemAgent},j}$ and $T_{\mathrm{method},j}$ denote the
inference time of MemAgent and the evaluated method on question $j$.
Speedup relative to MemAgent is
\[
\operatorname{Speedup}=
\frac{\sum_j T_{\mathrm{MemAgent},j}}{\sum_j T_{\mathrm{method},j}}.
\]
MemAgent has a speedup of $1\times$; values above $1\times$ indicate faster
inference than MemAgent.

\paragraph{Output tokens and reduction.}
Output tokens include System~1's output and System~2's verification,
memory generation, and final-answer output. For the same length setting and
question cohort, let $O_{\mathrm{MemAgent},j}$ and $O_{\mathrm{method},j}$
denote the respective output-token counts on question $j$. The output-token
ratio $r_O$ and percentage reduction relative to MemAgent are
\[
\begin{aligned}
r_O&=\frac{\sum_j O_{\mathrm{method},j}}{\sum_j O_{\mathrm{MemAgent},j}},\\
\operatorname{Reduction}(\%)&=100(1-r_O).
\end{aligned}
\]
MemAgent has $r_O=1$ and zero reduction; a positive reduction means fewer
output tokens than MemAgent.

\section{Training and Implementation Details}
\label{app:training-implementation}

We train MemDuet on 8~$\times$~NVIDIA A100 (80GB) GPUs.
For inference, we use 1~$\times$~NVIDIA A100 (80GB) GPU
and process one example at a time (batch size 1).

\subsection{Stage I: Memory-Contrastive Supervised Warm-up}
\label{app:training-stage-one}
\label{app:training-configuration}

\paragraph{Training data.}

We use 32,758 HotpotQA training questions \citep{yang2018hotpotqa} with
complete RL-MemAgent-7B teacher-memory trajectories. We randomly select
500 questions for validation and use the remaining 32,258 questions for
training. All chunks and memory pairs from the same question
remain in the same split.

Contexts are divided into consecutive 5,000-token chunks with the teacher
tokenizer. The first chunk initializes memory and is excluded from action
training examples. Gold supporting facts identify evidence chunks, and
the teacher updates memory only at these chunks in document order. Each
subsequent chunk is paired with the question and the memory before that
chunk: evidence chunks yield UPDATE examples, while other chunks yield
natural SKIP examples. For every UPDATE example, replacing only the memory
with the teacher's post-update memory yields an additional SKIP example:
\begin{equation}
\begin{aligned}
(q,c,m)&\longrightarrow\update{},\\
(q,c,m')&\longrightarrow\skipaction{}.
\end{aligned}
\end{equation}
Changing memory alone changes the target to \skipaction{}.
The training set contains 195,111 examples: 41,180 UPDATE examples,
112,751 natural SKIP examples, and 41,180 post-update SKIP examples.
The validation set contains 2,996 examples: 627 UPDATE examples,
1,742 natural SKIP examples, and 627 post-update SKIP examples.

\paragraph{Configuration.}
For Stage~I, we train a System~1 LoRA adapter \citep{hu2022lora}
with rank $r=32$,
scaling parameter $\alpha_{\mathrm{LoRA}}=64$, and dropout 0.05.
We use the AdamW optimizer with learning rate $10^{-4}$, a cosine annealing
schedule, and a warm-up ratio of 0.03. The effective batch size is 64,
the precision is bf16, and the training budget is three epochs.
The UPDATE and SKIP class weights are 13 and 1, respectively; the
paired-memory hinge loss has margin 1 and weight 0.2.
The class-weighted action cross-entropy is normalized by the total
target-class weight in each batch.

\input{sections/appendix_training_stage_two}
\input{sections/appendix_training_stage_three}

\subsection{Complete MemDuet Inference}
\label{app:inference-configuration}
\label{app:inference-algorithm}

\input{sections/workflow_algorithm_en}

\input{sections/appendix_proprietary_api}

\input{sections/appendix_hotpot_results}

\section{Speculative Acceptance}
\label{app:speculative-acceptance}
\label{app:speculative-decoding}

This section distinguishes the coverage supplied by a high-recall selector
from the target-distribution guarantee of speculative decoding.

\subsection{Difference from Distribution-Preserving Speculative Decoding}

Standard speculative decoding combines acceptance with residual resampling
after rejection to preserve the target model's sampling distribution,
regardless of draft quality \citep{leviathan2023speculative,chen2023speculative}.
This is distributional equivalence, not a guarantee that the target answer
is correct. MemDuet instead speculates over memory-update actions.
A System~1 SKIP bypasses System~2, and a rejected UPDATE does not trigger
residual resampling. Thus Eq.~\eqref{eq:workflow-acceptance} alone does not
preserve System~2's standalone action or answer distribution.
Write $d_{t,i}$ and $v_{t,i}$ for the System~1 and System~2 UPDATE
probabilities, respectively, at the same $(q,c_i,m_t)$.
With sampled proposals, the probability of selecting and accepting a chunk is
$d_{t,i}\min(1,v_{t,i}/d_{t,i})=\min(d_{t,i},v_{t,i})$.
For $d_{t,i}=0.4$ and $v_{t,i}=0.6$, it is $0.4$,
below the verifier's standalone UPDATE probability $0.6$.

\subsection{Quantified Coverage under High Recall}

On a specified distribution of live, memory-conditioned states, let
$z=\update{}$ denote a reference UPDATE label. Define selector recall
$r=\Pr(i\in\mathcal I_t\mid z=\update{})$ and the mean acceptance probability
$\bar\alpha=\mathbb E[\alpha_{t,i}\mid i\in\mathcal I_t,z=\update{}]$.
If $r\geq1-\varepsilon$ and $\bar\alpha\geq1-\eta$, the independent uniform
acceptance test gives
\begin{equation}
\begin{aligned}
&\Pr(i\in\mathcal I_t,\xi_{t,i}\leq\alpha_{t,i}\mid z=\update{})\\
&\quad=r\bar\alpha\geq(1-\varepsilon)(1-\eta).
\end{aligned}
\label{eq:acceptance-recall-coverage}
\end{equation}
A sufficient condition for the acceptance bound is
$v_{t,i}\geq1-\eta$ on selected reference-positive states:
for $0<d_{t,i}\leq1$, $\min(1,v_{t,i}/d_{t,i})\geq v_{t,i}$.
The verifier cannot recover a candidate omitted by System~1.
As a hypothetical example, recall $0.99$ and mean acceptance $0.99$
yield reference-positive action coverage $0.9801$, not answer accuracy
$0.9801$.

If these bounds hold conditional on the live prefix at each of at most
$L$ reference-positive decisions per question, a union bound gives
\begin{equation}
p_{\mathrm{miss}}\leq
\min\{1,L(\varepsilon+\eta-\varepsilon\eta)\},
\label{eq:acceptance-miss-bound}
\end{equation}
where $p_{\mathrm{miss}}$ is the probability that at least one such decision
is skipped or rejected. No independence between decisions is required.
With acceptance probability one on positives, this reduces to
$p_{\mathrm{miss}}\leq\min\{1,L\varepsilon\}$.
Here, live decisions exclude stale results discarded after an earlier rewrite.
A held-out aggregate recall does not by itself establish the required
conditional bounds on reached states.

\input{sections/appendix_prompts_en}

\input{sections/appendix_cases_en}

%% file: sections/appendix_training_stage_two.tex
\subsection{\apptext{Stage II: Optimizing System 2 with a Fixed Selector}}
\label{app:training-stage-two}

\subsubsection{\apptext{Dataset Construction}}
\apptext{We use MemAgent's released long-context HotpotQA data
\citep{yu2026memagent} for the 32,768 training examples derived from
the HotpotQA training split. The 128 questions in MemAgent's released
development set are reserved exclusively for final RULER-HQA testing
across context lengths (Appendix~\ref{app:data-preprocessing}).
For periodic validation and checkpoint selection, we sample 128 additional
questions from the remaining HotpotQA validation split with random seed 42.
Their long contexts use the same evidence-plus-distractor construction.
The resulting validation set is fixed before training, and test results
are not used to select checkpoints. Each example pairs a question
$q$ with its long context $D$ and reference answer $y^\ast$:}
\begin{equation}
\mathcal D_{\mathrm{II}}=\{(q,D,y^\ast)\}.
\label{eq:appendix-stage-two-data}
\end{equation}
\apptext{For training, we apply center truncation to each long context
and divide it into consecutive 5,000-token chunks in document order,
retaining at most eight chunks per question. All chunks of an example
share the same question and reference answer.}

\apptext{The offline data contain no reference-memory sequence or fixed
candidate indices. System~2 generates initialization and replacement
memories online. The frozen selector is reevaluated using each generated
memory.}

\subsubsection{\apptext{Reward Function}}
\apptext{We compute answer rewards using normalized exact matching of
boxed answers. For each trajectory, we extract the answer inside the
last \texttt{\textbackslash boxed\{...\}} field in the final 300 output
characters as the prediction. We lowercase the prediction and reference
answer aliases and apply formatting normalization. The answer reward
is 1 if the prediction exactly matches any normalized reference alias;
missing, malformed, or unmatched answers receive 0.}

\apptext{For trajectory $g$, let $r_g^{\mathrm{ans}}$ denote the binary
score of the final answer $y_g$. We use this score as the training reward:}
\begin{equation}
R_g=R_{\mathrm{task}}(y_g,y^\ast)=r_g^{\mathrm{ans}}.
\label{eq:appendix-stage-two-recorded-reward}
\end{equation}
\apptext{For each question, we sample $G=16$
complete trajectories and compute standardized group-relative advantages
using Eq.~\eqref{eq:stage-two-advantage}. Each trajectory's advantage is
shared by its verification commitment events and all generated tokens
for System~2 policy optimization.}

\subsubsection{\apptext{Training Configuration}}
\apptext{For the full configuration, we initialize System~1 from the Stage~I Qwen3.5-2B model
and freeze the selector throughout Stage~II.
System~2 is initialized from Qwen2.5-7B-Instruct,
and all its parameters are optimized.
For KL regularization, the generation reference policy is a frozen copy
of this initial Qwen2.5-7B-Instruct model. It remains fixed throughout Stage~II,
while the behavior snapshot $\theta_{\mathrm{old}}$ is refreshed each iteration.}

\apptext{Stage~II collects complete trajectories under
$(\phi_{\mathrm{SFT}},\theta_{\mathrm{old}})$.
System~2 initializes memory from the first chunk, and the frozen selector
evaluates a four-chunk window under the current memory. System~2 scores
the selected candidates; independent acceptance tests follow
Eq.~\eqref{eq:workflow-acceptance}. Only the first accepted candidate in
document order triggers a memory rewrite. Subsequent decisions under the
old memory are discarded, and processing resumes after the committed chunk.
If no candidate is accepted, memory is unchanged and the cursor advances
past the window. After document processing, System~2 generates the final
answer from the question and final memory.}

\apptext{We use AdamW to update all System~2 parameters with a learning
rate of $10^{-6}$. The selector uses a four-chunk window.
Table~\ref{tab:stage-two-configuration} summarizes the configuration.}

\begin{table}[!htbp]
\centering
\fontsize{8.5}{10}\selectfont
\caption{\apptext{Stage II training configuration.}}
\label{tab:stage-two-configuration}
\setlength{\tabcolsep}{3pt}
\begin{tabularx}{\linewidth}{@{}>{\raggedright\arraybackslash}X>{\raggedright\arraybackslash}X@{}}
\toprule
\apptext{Setting} & \apptext{Value}\\
\midrule
\apptext{Training / validation instances}
& 32,768 / 128\\
\apptext{Questions $\times$ rollouts}
& $128\times16=2{,}048$\\
\apptext{Question limit}
& 1,024 \apptext{tokens}\\
\apptext{Chunk / memory / answer limits}
& 5,000 / 1,024 / 1,024 \apptext{tokens}\\
\apptext{Maximum chunks / selector window}
& 8 / 4\\
\apptext{Trainable parameters}
& \apptext{All System~2; System~1 frozen}\\
\apptext{Optimizer / learning rate}
& AdamW / $10^{-6}$\\
\apptext{Weight decay / gradient clipping}
& 0.01 / 1.0\\
\apptext{Warm-up / schedule}
& \apptext{20 steps; then constant}\\
\apptext{PPO question minibatch / passes}
& 8 / 1\\
\apptext{Policy clip}
& 0.2\\
\apptext{Generation KL / entropy coefficient}
& $10^{-3}$ / 0\\
\apptext{Training temperature / top-$p$}
& 1 / 1\\
\apptext{Validation rollouts / temperature / top-$p$}
& 4 / 1 / 0.7\\
\apptext{Training budget}
& 25 epochs\\
\apptext{Save / validation interval}
& \apptext{10 / 5 rollout batches}\\
\apptext{Rollout precision}
& bfloat16\\
\bottomrule
\end{tabularx}
\end{table}

\apptext{With a training budget of 25 epochs, Stage~II minimizes
Eq.~\eqref{eq:stage-two-objective}. Clipped policy losses are averaged
over $\mathcal E_g$ separately within each trajectory. KL terms are
averaged only over generated tokens $\mathcal T_g$, with
$\beta=10^{-3}$; verification commitment events receive no KL penalty.
Commitment probabilities follow the controller-induced distribution
described after Eq.~\eqref{eq:stage-two-ratio}.}
\apptext{For $n_t=|\mathcal I_t|>0$, the verification event
$e=(\mathrm{ver},t)$ records a global chunk index $i_{g,t}^\star$ or
$0$, conditioned on $(q,\mathcal W_{g,t},\mathcal I_{g,t},m_{g,t})$.
This is one verification event per window, marginalizing
outcomes after the earliest accepted candidate. An empty candidate set
is deterministic and contributes no verification-policy loss.
The verification commitment and ordinary memory-update ratios follow
Eqs.~\eqref{eq:stage-two-ratio}--\eqref{eq:stage-two-explicit-ratios}.
Initialization uses $(q,c_1,m_{\mathrm{init}},m_{g,0,<\ell})$ to score
$m_{g,0,\ell}$, and is included among trainable memory-generation tokens.
Final answer generation uses the token-level ratio with conditions
$(q,m_{g,\mathrm{fin}},y_{g,<\ell})$ defined in Stage~II; its tokens also
contribute to the policy loss and KL regularizer.
During replay, we update parameters under the fixed rollout conditions
and mask inputs and padding in generation losses. Identical-reward
groups have zero policy-advantage contribution, but the KL term remains.}

\subsubsection{\apptext{Training Pseudocode}}
\label{app:training-stage-two-pseudocode}

\apptext{Algorithm~\ref{alg:stage-two-training} details Stage~II full-parameter
optimization of System~2 with a fixed System~1.}

\begin{tcolorbox}[appendixpanel,breakable,colback=white,
step and label={algorithm}{alg:stage-two-training},
colframe=appendixink,colbacktitle=appendixink,coltitle=white,
title={\apptext{Algorithm~\thealgorithm: Stage II training with a fixed selector}}]
\fontsize{8.5}{10}\selectfont
\begin{algorithmic}[1]
\Require $\mathcal D_{\mathrm{II}}$, $\phi_{\mathrm{SFT}}$, initial $\theta$
\State Freeze $\phi\gets\phi_{\mathrm{SFT}}$
\State Fix $\pi_{\mathrm{ref}}^{\mathrm{gen}}$ as a frozen copy of initial System~2
\For{each training iteration}
  \State $\theta_{\mathrm{old}}\gets\operatorname{snapshot}(\theta)$
  \State Sample $N_{\mathrm{batch}}$ question--document pairs
  \For{each pair $(q,D,y^\ast)$}
    \For{$g=1,\ldots,G$}
      \State Execute the workflow with $(\phi,\theta_{\mathrm{old}})$
      \State Store inputs, events, behavior probabilities, and $R_g$
    \EndFor
    \State Compute advantages using Eq.~\eqref{eq:stage-two-advantage}
  \EndFor
  \State Zero gradients; replay each event under current $\theta$
  \State Form all event ratios from Section~\ref{sec:training-stage-two}
  \State Average clipped event losses per trajectory
  \State Add generated-token KL with $\beta=10^{-3}$
  \State Backpropagate Eq.~\eqref{eq:stage-two-objective}; clip gradients
  \State Update all parameters $\theta$; synchronize the rollout policy
  \State Discard rollouts; validate and save on schedule
\EndFor
\end{algorithmic}
\end{tcolorbox}

%% file: sections/appendix_training_stage_three.tex
\subsection{\apptext{Stage III: Joint Optimization of Selection and Consolidation}}
\label{app:training-stage-three}

\paragraph{\apptext{Complete trajectory distribution.}}
\apptext{For a legal controller trajectory, let $j_t=i_t^\star$ denote the
recorded commitment outcome. The selection policy, commitment distribution,
and System~2 generation processes jointly induce}
\input{sections/training_trajectory_distribution}
\apptext{Products include only actual rounds and generated tokens;
without a commitment, $m_{t+1}=m_t$. Verification enters through
$\alpha_{t,i}$ in the commitment probability from
the distribution described after Eq.~\eqref{eq:stage-two-ratio}, with no additional
$\pi_\theta^{\mathrm{ver}}$ factor. Thus $P_{\phi,\theta}$ is the
distribution induced by the complete workflow, not a separate policy.}

\subsubsection{\apptext{Dataset Construction}}
\apptext{For Stage~III, we reuse the Stage~II dataset in
Eq.~\eqref{eq:appendix-stage-two-data}, with
$\mathcal D_{\mathrm{III}}=\mathcal D_{\mathrm{II}}$.
The training set retains the 32,768 examples, and checkpoint selection
uses the same separate 128-question validation set as Stage~II.
The 128 questions in MemAgent's released development set remain reserved
for final testing. No extra supervised action labels,
reference memories, or acceptance-derived correctness labels are introduced.
The change is online trajectory generation: both current policies produce
the selections, verification outcomes, intermediate memories, and answers.}

\subsubsection{\apptext{Reward Function}}
\apptext{Both policies reuse the Stage~II final-answer reward in
Eq.~\eqref{eq:appendix-stage-two-recorded-reward}. The group-relative advantage follows
Eq.~\eqref{eq:stage-two-advantage}; a single trajectory's advantage is
assigned to selection, verification, memory-generation, and answer events.
Accepting a candidate is not an independent correctness reward, and
a \skipaction{} decision is not omitted merely because it avoids a
large-model generation call.}

\subsubsection{\apptext{Training Configuration}}
\apptext{For the full configuration, System~1 uses Qwen3.5-2B and
continues training the LoRA adapters from the selected Stage~I checkpoint.
System~2 loads the selected Qwen2.5-7B-Instruct checkpoint
from full-parameter Stage~II training and adds new LoRA adapters.
Both backbones remain frozen; optimization
updates only the adapter parameters within $\phi$ and $\theta$.
The two systems use separate AdamW optimizers with independent optimizer states
and synchronized behavior snapshots $(\phi_{\mathrm{old}},\theta_{\mathrm{old}})$.
Optimizer and scheduler states are reset at the start of Stage~III,
while checkpoint weights are retained. Dropout is disabled during both
rollout and policy replay. Table~\ref{tab:stage-three-configuration}
summarizes the configuration.}

\begin{table*}[!t]
\centering
\fontsize{8.5}{10.5}\selectfont
\caption{\apptext{Stage III joint LoRA training configuration.}}
\label{tab:stage-three-configuration}
\setlength{\tabcolsep}{4pt}
\renewcommand{\arraystretch}{1.12}
\begin{tabularx}{\textwidth}{@{}>{\raggedright\arraybackslash}p{0.25\textwidth}>{\raggedright\arraybackslash}X>{\raggedright\arraybackslash}X@{}}
\toprule
\apptext{Setting} & \apptext{System 1: selector}
& \apptext{System 2: verifier / memory updater}\\
\midrule
\apptext{Backbone} & Qwen3.5-2B & Qwen2.5-7B-Instruct\\
\apptext{Trainable parameters}
& \apptext{Existing LoRA adapters; backbone frozen}
& \apptext{New LoRA adapters; backbone frozen}\\
\apptext{LoRA rank / scaling} $(r/\alpha_{\mathrm{LoRA}})$
& 32 / 64 & 32 / 32\\
\apptext{LoRA target modules}
& \apptext{Same as Stage~I}
& All-linear\\
\apptext{Dropout during joint RL}
& \apptext{Disabled in rollout and policy replay}
& \apptext{Disabled in rollout and policy replay}\\
\midrule
\apptext{Training budget}
& \multicolumn{2}{l}{\apptext{5 dataset epochs}}\\
\apptext{Global rollout batch}
& \multicolumn{2}{l}{\apptext{128 questions; $G=16$ rollouts per question}}\\
\apptext{Optimization minibatch}
& \multicolumn{2}{l}{\apptext{8 questions, retaining all 16 rollouts per question}}\\
\apptext{PPO passes}
& \multicolumn{2}{l}{\apptext{1 per rollout batch}}\\
\apptext{Training sampling}
& \multicolumn{2}{l}{temperature $=1.0$, top-$p=1.0$}\\
\apptext{Validation sampling}
& \multicolumn{2}{l}{$G=4$, temperature $=1.0$, top-$p=0.7$}\\
\apptext{Policy-ratio clip $\epsilon$}
& \multicolumn{2}{l}{0.2}\\
\midrule
\apptext{Optimizer}
& \apptext{AdamW; independent state}
& \apptext{AdamW; independent state}\\
\apptext{Peak learning rate} & $10^{-6}$ & $5\times10^{-7}$\\
Adam $(\beta_1,\beta_2,\epsilon)$
& $(0.9,0.999,10^{-8})$ & $(0.9,0.999,10^{-8})$\\
\apptext{Weight decay} & 0.01 & 0.01\\
\apptext{Maximum gradient norm}
& \apptext{1.0; clipped separately}
& \apptext{1.0; clipped separately}\\
\apptext{Learning-rate schedule}
& \apptext{20 optimizer steps of warm-up, then constant}
& \apptext{20 optimizer steps of warm-up, then constant}\\
\apptext{Optimizer / scheduler initialization}
& \apptext{Reset at Stage~III start}
& \apptext{Reset at Stage~III start}\\
\midrule
\apptext{Policy-gradient sources}
& \apptext{Selection and joint verification commitments}
& \apptext{Verification commitments; initialization, memory-update, and answer tokens}\\
\apptext{Generation KL coefficient}
& \apptext{Not applicable} & $\beta=10^{-3}$\\
\apptext{Generation reference policy}
& \apptext{Not applicable}
& \apptext{Same frozen Qwen2.5-7B-Instruct reference as Stage~II}\\
\bottomrule
\end{tabularx}
\end{table*}

\apptext{Selection events update only System~1 adapters using the ratio in
Eq.~\eqref{eq:stage-three-event-ratios}, with the recorded action
$a_{g,t,i}$ and conditions $(q,c_i,m_{g,t})$.
Verification commitments update both systems' adapters through}
\begin{equation}
\rho_{g,t}^{\mathrm{ver}}=
\frac{\Pr_{\phi,\theta}
(i_t^\star=i_{g,t}^\star\mid q,\mathcal W_{g,t},\mathcal I_{g,t},m_{g,t})}
{\Pr_{\phi_{\mathrm{old}},\theta_{\mathrm{old}}}
(i_t^\star=i_{g,t}^\star\mid q,\mathcal W_{g,t},\mathcal I_{g,t},m_{g,t})},
\label{eq:appendix-joint-verification-ratio}
\end{equation}
\apptext{where $i_{g,t}^\star$ is the recorded committed chunk index or
$0$. The full conditions are the recorded query, window,
candidate indices, and memory. The commitment probabilities follow
the distribution described after Eq.~\eqref{eq:stage-two-ratio}; they retain
dependence on both systems through the acceptance ratio. Behavior
probabilities are fixed. Candidate sets, chunk indices, action labels,
and memory strings are replayed as conditioning data, without regenerating
them or differentiating through discrete text. Ordinary memory-update
tokens use the System~2 ratio in Eq.~\eqref{eq:stage-two-explicit-ratios};
initialization and final answer generation use the substitutions and
full conditions defined in Stage~II. Their policy loss and KL regularizer
update only System~2 adapters. The complete trajectory factors
follow Eq.~\eqref{eq:workflow-trajectory-distribution}, with no additional
verifier-policy factor.}

\apptext{Let $\mathcal E_g$ contain all four event types.
The clipped loss averages events within each trajectory, as in Stage~II.
Generation-only KL retains $\beta=10^{-3}$ and the same frozen reference
policy as Stage~II: a frozen copy of the initial Qwen2.5-7B-Instruct model.
This reference remains fixed across both RL stages. Selection and
verification events receive no KL term. Both adapters' gradients
are computed before either optimizer is stepped, and both rollout
policies are synchronized before collecting a fresh group.}

\subsubsection{\apptext{Training Pseudocode}}
\label{app:training-stage-three-pseudocode}

\apptext{Algorithm~\ref{alg:stage-three-training} summarizes joint policy optimization.}

\begin{tcolorbox}[appendixpanel,breakable,colback=white,
step and label={algorithm}{alg:stage-three-training},
colframe=appendixink,colbacktitle=appendixink,coltitle=white,
title={\apptext{Algorithm~\thealgorithm: Stage III joint policy optimization}}]
\fontsize{8.5}{10}\selectfont
\begin{algorithmic}[1]
\Require $\mathcal D_{\mathrm{III}}$, $\phi_{\mathrm{SFT}}$, $\theta_{\mathrm{II}}$;
Stage~II reference $\pi_{\mathrm{ref}}^{\mathrm{gen}}$
\State Load both checkpoints; freeze both backbones
\State Resume System~1 LoRA; add new System~2 LoRA
\State Retain the frozen reference; disable dropout
\State Initialize separate AdamW optimizers and schedules
\For{each joint-training iteration}
  \State Snapshot $(\phi_{\mathrm{old}},\theta_{\mathrm{old}})$
  \State Sample $N_{\mathrm{batch}}$ question--document pairs
  \For{each pair $(q,D,y^\ast)$}
    \State Collect $G$ fresh complete rollouts using the snapshots
    \State Record every selection and each nonempty-window verification
    \State Record initialization, memory, and answer tokens
    \State Compute rewards using Eq.~\eqref{eq:appendix-stage-two-recorded-reward}
    and advantages using Eq.~\eqref{eq:stage-two-advantage}
  \EndFor
  \For{\apptext{each 8-question minibatch (all $G$ rollouts/question)}}
    \State Zero both adapters' gradients
    \State Replay actions under $\phi$ and tokens under $\theta$
    \State Score commitments using the distribution after Eq.~\eqref{eq:stage-two-ratio}
    \State Average clipped losses over $\mathcal E_g$ per trajectory
    \State Add System~2 generated-token KL with $\beta=10^{-3}$
    \State Backpropagate before stepping either optimizer
    \State Clip adapter gradients separately; update both adapters
  \EndFor
  \State Synchronize both rollout models; discard old rollouts
  \State Validate and save the current pair on schedule
\EndFor
\end{algorithmic}
\end{tcolorbox}

%% file: sections/training_trajectory_distribution.tex
\begin{equation}
\begin{aligned}
&P_{\phi,\theta}(\tau\mid q,D)\\
&=\pi_\theta^{\mathrm{mem}}(m_0\mid q,c_1,m_{\mathrm{init}})\\
&\quad\times\prod\nolimits_t\Bigl[
\prod\nolimits_{i:\,c_i\in\mathcal W_t}
\pi_\phi^{\mathrm{sel}}(a_{t,i}\mid q,c_i,m_t)\\
&\qquad\times\Pr_{\phi,\theta}
(i_t^\star=j_t\mid q,\mathcal W_t,\mathcal I_t,m_t)\Bigr]\\
&\quad\times\prod\nolimits_{t:\,j_t\ne0}
\pi_\theta^{\mathrm{mem}}(m_{t+1}\mid q,c_{j_t},m_t)\\
&\quad\times\prod\nolimits_{\ell=1}^{|y|}
\pi_\theta^{\mathrm{gen}}(y_\ell\mid q,m_{\mathrm{fin}},y_{<\ell}).
\end{aligned}
\label{eq:workflow-trajectory-distribution}
\end{equation}

%% file: sections/workflow_algorithm_en.tex
\begin{tcolorbox}[appendixpanel,breakable,colback=white,
step and label={algorithm}{alg:memduet-round},
colframe=appendixink,colbacktitle=appendixink,coltitle=white,
title={Algorithm~\thealgorithm: Complete MemDuet inference}]
\fontsize{8.5}{10}\selectfont
\begin{algorithmic}[1]
\Require Query $q$, nonempty ordered chunks $(c_1,\ldots,c_T)$,
window width $k$
\Ensure Final answer $y$
\State $m_{\mathrm{init}}\gets\texttt{No previous memory}$
\State Generate $m_0$ under $\pi_\theta^{\mathrm{mem}}(\cdot\mid q,c_1,m_{\mathrm{init}})$
\State $t\gets0$; $b_0\gets2$
\While{$b_t\leq T$}
  \State $h_t\gets\min(b_t+k-1,T)$
  \State $\mathcal W_t\gets\{c_{b_t},\ldots,c_{h_t}\}$; fix memory $m_t$
  \State In parallel, obtain System~1 scores for
  each $(q,c_i,m_t)$, $i=b_t,\ldots,h_t$
  \State $d_{t,i}\gets
  \frac{e^{z^s_{u,t,i}}}{e^{z^s_{k,t,i}}+e^{z^s_{u,t,i}}}$
  \State Independently sample $a_{t,i}\sim\pi_\phi^{\mathrm{sel}}(\cdot\mid q,c_i,m_t)$
  for $i=b_t,\ldots,h_t$
  \State $\mathcal I_t\gets\{i:c_i\in\mathcal W_t,\ a_{t,i}=\update{}\}$
  \State For $i\in\mathcal I_t$, obtain System~2 scores
  for the same $(q,c_i,m_t)$
  \State $v_{t,i}\gets
  \frac{e^{z^l_{u,t,i}}}{e^{z^l_{k,t,i}}+e^{z^l_{u,t,i}}}$
  \State For $i\in\mathcal I_t$, set $\alpha_{t,i}\gets\min(1,v_{t,i}/d_{t,i})$
  \State Record independent $\xi_{t,i}\sim\mathcal U(0,1)$ for $i\in\mathcal I_t$
  \If{no $i\in\mathcal I_t$ satisfies $\xi_{t,i}\leq\alpha_{t,i}$}
    \State $i_t^\star\gets0$; $m_{t+1}\gets m_t$
    \State $b_{t+1}\gets h_t+1$ \Comment{all skipped or rejected}
  \Else
    \State $i_t^\star\gets\min\{i\in\mathcal I_t\mid\xi_{t,i}\leq\alpha_{t,i}\}$
    \State Generate $m_{t+1}$ under $\pi_\theta^{\mathrm{mem}}(\cdot\mid q,c_{i_t^\star},m_t)$
    \State $b_{t+1}\gets i_t^\star+1$
    \State Discard every later result computed from $m_t$
  \EndIf
  \State $t\gets t+1$
\EndWhile
\State Generate $y$ autoregressively under
$\pi_\theta^{\mathrm{gen}}(\cdot\mid q,m_t,y_{<\ell})$
\State \Return $y$
\end{algorithmic}
\end{tcolorbox}

%% file: sections/appendix_proprietary_api.tex
\subsection{\apptext{Training-Free Transfer to Proprietary APIs}}
\label{app:proprietary-api}

\paragraph{\apptext{Models and API versions.}}
\apptext{
For DeepSeek, we use
DeepSeek-V4.1-Flash as System~1 and DeepSeek-V4-Pro-0813 as System~2.
For OpenAI, we use GPT-5.6 Luna as System~1 and GPT-5.6 Sol as System~2.
Transfer uses task prompts and workflow control without additional SFT
or GRPO.
}

\paragraph{\apptext{Differences from the full MemDuet method.}}
\apptext{
The API adaptation differs primarily in training and acceptance.
The full method learns the two systems through the three-stage training
in Section~\ref{sec:smpi}. Here, we instead use off-the-shelf proprietary
models with fixed parameters, adapting their roles through task prompts
and an external controller rather than additional SFT or GRPO.
}

\apptext{
In the full method, the UPDATE probabilities of System~1 and System~2 determine
the acceptance probability in Eq.~\ref{eq:workflow-acceptance}, followed
by independent random acceptance tests. In the API adaptation, we do not
request action log-probabilities. Instead, generated \texttt{k}/\texttt{u}
labels drive the decisions: System~1's \texttt{k} skips a chunk, while
\texttt{u} forwards it to System~2; System~2's \texttt{k} rejects the
candidate, while \texttt{u} directly accepts it. We therefore replace
probability-ratio-based speculative acceptance with text verification,
without computing the acceptance ratio or drawing separate acceptance
random variables.
}

\paragraph{\apptext{Request settings.}}
\apptext{
All calls to both systems use thinking-disabled mode.
The output caps are 8,192 tokens for System~1 and 32,768 for System~2.
Contexts are split into 5,000-token
chunks, with a four-chunk lookahead and a 1,024-token memory budget.
}

\paragraph{\apptext{API prices.}}
\apptext{
Table~\ref{tab:api-rate-settings} lists prices in USD per million tokens.
The DeepSeek entries are \emph{off-peak} rates; peak rates are twice these
values.
}

\begin{table}[!htbp]
\centering
\fontsize{8.5}{10}\selectfont
\setlength{\tabcolsep}{3pt}
\begin{tabularx}{\linewidth}{@{}>{\raggedright\arraybackslash}Xrrr@{}}
\toprule
\apptext{Model} & \apptext{Input} & \apptext{Cached input} & \apptext{Output}\\
\midrule
DeepSeek-V4.1-Flash & 0.15 & 0.003 & 0.60\\
DeepSeek-V4-Pro-0813 & 0.66 & 0.022 & 1.98\\
GPT-5.6 Luna & 0.20 & 0.020 & 1.20\\
GPT-5.6 Sol & 4.00 & 0.400 & 20.00\\
\bottomrule
\end{tabularx}
\caption{\apptext{API prices in USD per million tokens. DeepSeek prices are off-peak rates.}}
\label{tab:api-rate-settings}
\end{table}

%% file: sections/appendix_hotpot_results.tex
\section{\apptext{Complete Experimental Results}}
\label{app:complete-results}

\subsection{\apptext{RULER-HQA Results}}
\label{app:complete-ruler-results}

\apptext{Table~\ref{tab:discussion-length-scaling} presents the complete
RULER-HQA results across different context lengths, reporting answer
accuracy, time speedup, and relative output-token counts for the compared
methods. Both efficiency metrics use MemAgent-7B as the baseline.}

\apptext{\input{sections/table1_integrated_en}}
\apptext{}
\input{sections/appendix_s1_backbone_results}

%% file: sections/table1_integrated_en.tex
\begingroup
\newcommand{\tableoneBeginFloat}{\begin{table*}[p]}
\newcommand{\tableoneMemDuetMarker}{}
\newcommand{\tableoneEstimateMarker}{}
\newcommand{\tableoneReproducedAcc}[1]{\accnum{#1}$^{*}$}
\newcommand{\tableoneLycheeNoGateLastAccuracy}{\nodata}
\input{sections/table1_qwen_instruct_accuracy_rows}
\newcommand{\tableoneTokens}{\textit{Efficiency}: Output tokens (MemAgent = $1\times$, $\downarrow$)}
\input{sections/table1_total_output_token_cells}
\newcommand{\tableoneMemDuetTokenCells}{\tableoneTotalOutputCells}
\definecolor{oursrow}{HTML}{E7E7E7}
\definecolor{metricblue}{HTML}{0044CC}
\let\tableoneoriginalcaption\caption
\renewcommand{\caption}[1]{\tableoneoriginalcaption{RULER-HQA length scaling.
The three panels report accuracy (\%; local values use sub-EM), time speedup
relative to MemAgent-7B, and relative output tokens (MemAgent = $1\times$).
Output tokens include both systems' outputs.
$^{*}$ denotes results from our reproduction.
Within each column and panel, \textbf{bold} and \underline{underlined} values indicate
the best and second-best distinct values, respectively.
``-'' denotes unreported results; N/A denotes unavailable results.}}
\input{sections/table1_discussion_horizontal}
\endgroup

%% file: sections/table1_qwen_instruct_accuracy_rows.tex
\newcommand{\tableoneExtraQwenAccuracyRows}{%
\modelname{Qwen2.5-Instruct}{32B} & \accnum{69.53} & \accnum{64.84} & \accnum{60.16} & \accnum{51.56} & \accnum{44.53} & \accnum{21.88} & \accnum{14.06} & \accnum{7.03} & \accnum{N/A} & \accnum{N/A} \\
\modelname{Qwen2.5-Instruct}{14B} & \accnum{75.00} & \accnum{67.19} & \accnum{57.03} & \accnum{54.69} & \accnum{44.53} & \accnum{21.88} & \accnum{10.94} & \accnum{2.34} & \accnum{N/A} & \accnum{N/A} \\
\modelname{Qwen2.5-Instruct}{7B} & \accnum{52.34} & \accnum{57.03} & \accnum{51.56} & \accnum{44.53} & \accnum{32.81} & \accnum{13.28} & \accnum{6.25} & \accnum{1.56} & \accnum{N/A} & \accnum{N/A} \\
}

%% file: sections/table1_discussion_horizontal.tex
\providecommand{\tableoneMethod}{Method}
\providecommand{\tableoneMemDuetMarker}{$^{\ddagger}$}
\providecommand{\tableoneEstimateMarker}{$^{\dagger}$}
\providecommand{\tableoneLycheeNoGateLastAccuracy}{\nodata}
\providecommand{\tableoneReproducedAcc}[1]{\accnum{#1}}
\providecommand{\tableoneExtraQwenAccuracyRows}{}
\providecommand{\tableonePerformance}{\textit{Performance}: Accuracy (\%, $\uparrow$)}
\providecommand{\tableoneSpeed}{\textit{Efficiency}: Time speedup ($\times$, $\uparrow$)}
\providecommand{\tableoneTokens}{\textit{Efficiency}: Relative updater output tokens (MemAgent = $1\times$, $\downarrow$)}
\providecommand{\tableoneMemDuetTokenCells}{\oursratio{\best{0.9023}} & \oursratio{\best{0.5622}} & \oursratio{\best{0.3624}} & \oursratio{\best{0.1975}} & \oursratio{\best{0.1125}} & \oursratio{\best{0.0585}} & \oursratio{\best{0.0525}} & \oursratio{\best{0.0444}} & \oursratio{\best{0.0103}} & \oursratio{\best{0.0075}}}
\providecommand{\tableoneBeginFloat}{\begin{table*}[!t]}
\tableoneBeginFloat
\centering
\begingroup
\definecolor{tableoneblue}{HTML}{214E97}
\definecolor{tableoneband}{HTML}{F1F3F7}
\definecolor{tableoneours}{HTML}{EAF1FA}
\definecolor{tableonetext}{HTML}{222222}
\definecolor{tableoneoutline}{HTML}{6C757E}
\color{tableonetext}
\fontsize{10}{12}\selectfont
\setlength{\tabcolsep}{0.7pt}
\renewcommand{\arraystretch}{1.21}
\newcommand{\nodata}{\textemdash}
\newcommand{\accnum}[1]{{\fontsize{8.5}{10}\selectfont #1}}
\newcommand{\modelname}[2]{\mbox{#1\kern0.8pt\raisebox{-0.45ex}{\fontsize{8}{9}\selectfont #2}}}
\newcommand{\best}[1]{\textbf{#1}}
\newcommand{\runnerup}[1]{\underline{#1}}
\newcommand{\speednum}[1]{{\fontsize{8.5}{10}\selectfont #1$\times$}}
\newcommand{\speedest}[1]{{\fontsize{8.5}{10}\selectfont #1$\times$}}
\newcommand{\ratioval}[1]{{\fontsize{8.5}{10}\selectfont #1$\times$}}
\newcommand{\oursacc}[1]{\textcolor{tableoneblue}{\accnum{#1}}}
\newcommand{\oursspeed}[1]{\textcolor{tableoneblue}{\speednum{#1}}}
\newcommand{\oursratio}[1]{\textcolor{tableoneblue}{\ratioval{#1}}}
\begin{tabularx}{\textwidth}{@{}l*{10}{>{\centering\arraybackslash}X}@{}}
\arrayrulecolor{tableoneoutline}\toprule
\textbf{\tableoneMethod} & \textbf{7K} & \textbf{14K} & \textbf{28K} & \textbf{56K} & \textbf{112K} & \textbf{224K} & \textbf{448K} & \textbf{896K} & \textbf{1.75M} & \textbf{3.5M} \\
\arrayrulecolor{black}\midrule
\rowcolor{tableoneband}\multicolumn{11}{c}{\textcolor{tableoneblue}{\itshape\tableonePerformance}} \\
\midrule
\modelname{RAG}{7B} & \accnum{67.19} & \accnum{66.41} & \accnum{66.41} & \accnum{67.19} & \accnum{64.84} & \accnum{64.06} & \accnum{62.50} & \accnum{61.72} & \accnum{62.38} & \nodata \\
\modelname{Search-R1}{7B} & \accnum{72.66} & \accnum{71.88} & \accnum{67.71} & \accnum{73.96} & \accnum{66.67} & \accnum{62.50} & \accnum{64.58} & \accnum{67.71} & \accnum{67.19} & \nodata \\
\midrule
\modelname{QwenLong-L1}{32B} & \accnum{72.66} & \accnum{75.00} & \accnum{72.66} & \accnum{60.94} & \accnum{31.25} & \accnum{17.19} & \accnum{13.28} & \accnum{11.72} & \accnum{N/A} & \accnum{N/A} \\
\shortstack[l]{QwenLong-L1.5\\[-2pt]{\fontsize{8}{9}\selectfont 30B-A3B}} & \accnum{67.97} & \accnum{68.75} & \accnum{66.41} & \accnum{63.28} & \accnum{61.72} & \accnum{59.38} & \accnum{43.75} & \accnum{39.06} & \accnum{28.13} & \nodata \\
\tableoneExtraQwenAccuracyRows
\modelname{Qwen2.5-1M}{14B} & \accnum{60.16} & \accnum{60.94} & \accnum{50.00} & \accnum{57.03} & \accnum{50.00} & \accnum{37.50} & \accnum{8.59} & \accnum{0.00} & \accnum{N/A} & \accnum{N/A} \\
\modelname{Qwen2.5-1M}{7B} & \accnum{61.72} & \accnum{56.25} & \accnum{53.91} & \accnum{55.47} & \accnum{51.56} & \accnum{33.59} & \accnum{12.50} & \accnum{0.00} & \accnum{N/A} & \accnum{N/A} \\
\midrule
\modelname{DS-Distill}{32B} & \accnum{70.31} & \accnum{66.41} & \accnum{65.62} & \accnum{46.88} & \accnum{23.44} & \accnum{13.28} & \accnum{7.81} & \accnum{7.03} & \accnum{N/A} & \accnum{N/A} \\
\modelname{DS-Distill}{14B} & \accnum{64.06} & \accnum{64.84} & \accnum{57.03} & \accnum{40.62} & \accnum{14.84} & \accnum{8.59} & \accnum{3.12} & \accnum{6.25} & \accnum{N/A} & \accnum{N/A} \\
\modelname{DS-Distill}{7B} & \accnum{30.47} & \accnum{12.50} & \accnum{3.12} & \accnum{0.00} & \accnum{0.00} & \accnum{0.78} & \accnum{0.00} & \accnum{0.00} & \accnum{N/A} & \accnum{N/A} \\
\midrule
\modelname{RL-MemAgent}{14B} & \accnum{80.47} & \accnum{82.03} & \accnum{\runnerup{82.03}} & \accnum{\best{83.59}} & \accnum{81.25} & \accnum{77.34} & \accnum{79.69} & \accnum{75.78} & \accnum{\runnerup{78.91}} & \accnum{71.09} \\
\modelname{RL-MemAgent}{7B} & \accnum{81.25} & \accnum{81.25} & \accnum{\runnerup{82.03}} & \accnum{80.47} & \accnum{79.69} & \accnum{75.78} & \accnum{76.56} & \accnum{74.22} & \accnum{77.34} & \accnum{71.88} \\
\modelname{InfMem}{4B} & \tableoneReproducedAcc{80.47} & \tableoneReproducedAcc{\runnerup{82.81}} & \tableoneReproducedAcc{72.66} & \tableoneReproducedAcc{75.78} & \tableoneReproducedAcc{76.56} & \tableoneReproducedAcc{75.00} & \tableoneReproducedAcc{65.62} & \tableoneReproducedAcc{\best{84.38}} & \tableoneReproducedAcc{78.12} & \tableoneReproducedAcc{68.75} \\
\modelname{Lychee ($-G$)}{7B} & \accnum{80.47} & \accnum{81.25} & \accnum{\runnerup{82.03}} & \accnum{\runnerup{81.25}} & \accnum{80.47} & \accnum{\runnerup{79.69}} & \accnum{75.00} & \accnum{75.78} & \accnum{78.12} & \tableoneLycheeNoGateLastAccuracy \\
\modelname{Lychee ($+G$)\tableoneEstimateMarker}{7B} & \accnum{77.34} & \accnum{76.56} & \accnum{75.00} & \accnum{76.56} & \accnum{75.78} & \accnum{73.44} & \accnum{74.22} & \accnum{72.66} & \accnum{71.09} & \nodata \\
\modelname{ReMemR1}{7B} & \tableoneReproducedAcc{\runnerup{82.03}} & \tableoneReproducedAcc{\runnerup{82.81}} & \tableoneReproducedAcc{81.25} & \tableoneReproducedAcc{78.91} & \tableoneReproducedAcc{\runnerup{82.81}} & \tableoneReproducedAcc{\runnerup{79.69}} & \tableoneReproducedAcc{\runnerup{80.47}} & \tableoneReproducedAcc{77.34} & \tableoneReproducedAcc{71.09} & \tableoneReproducedAcc{\runnerup{72.66}} \\
\rowcolor{tableoneours}\textcolor{tableoneblue}{\modelname{\textbf{MemDuet}\tableoneMemDuetMarker}{7B}} & \oursacc{\best{87.50}} & \oursacc{\best{83.59}} & \oursacc{\best{89.06}} & \oursacc{79.69} & \oursacc{\best{87.50}} & \oursacc{\best{85.94}} & \oursacc{\best{82.81}} & \oursacc{\runnerup{82.03}} & \oursacc{\best{79.69}} & \oursacc{\best{75.78}} \\
\midrule
\rowcolor{tableoneband}\multicolumn{11}{c}{\textcolor{tableoneblue}{\itshape\tableoneSpeed}} \\
\midrule
\modelname{MemAgent}{7B} & \speednum{\runnerup{1.0}} & \speednum{\runnerup{1.0}} & \speednum{1.0} & \speednum{1.0} & \speednum{1.0} & \speednum{1.0} & \speednum{1.0} & \speednum{1.0} & \speednum{1.0} & \speednum{1.0} \\
\modelname{InfMem}{4B} & \speedest{0.14} & \speedest{0.15} & \speednum{0.21} & \speedest{0.51} & \speednum{0.74} & \speedest{1.16} & \speedest{1.83} & \speedest{6.68} & \speedest{\runnerup{12.6}} & \speedest{\runnerup{6.11}} \\
\modelname{GRU-Mem\tableoneEstimateMarker}{7B} & \speedest{\best{1.09}} & \speedest{0.98} & \speedest{\runnerup{1.21}} & \speedest{\runnerup{1.57}} & \speedest{\runnerup{1.66}} & \speedest{2.04} & \speedest{2.30} & \speedest{2.40} & \nodata & \nodata \\
\modelname{Lychee ($-G$)}{7B} & \speedest{0.191} & \speedest{0.206} & \speednum{0.233} & \speedest{0.305} & \speednum{0.377} & \speedest{0.445} & \speedest{0.393} & \speedest{0.515} & \speedest{0.411} & \speedest{0.399} \\
\modelname{Lychee ($+G$)\tableoneEstimateMarker}{7B} & \speedest{0.19} & \speedest{0.25} & \speedest{0.37} & \speedest{0.76} & \speedest{1.32} & \speedest{\runnerup{2.62}} & \speedest{\runnerup{3.81}} & \speedest{\runnerup{9.12}} & \speedest{11.6} & \nodata \\
\modelname{ReMemR1}{7B} & \speedest{0.173} & \speedest{0.185} & \speedest{0.213} & \speedest{0.272} & \speedest{0.325} & \speedest{0.353} & \speedest{0.413} & \speedest{0.469} & \speedest{0.456} & \speedest{0.560} \\
\rowcolor{tableoneours}\textcolor{tableoneblue}{\modelname{\textbf{MemDuet}\tableoneMemDuetMarker}{7B}} & \oursspeed{0.983} & \oursspeed{\best{1.445}} & \oursspeed{\best{1.588}} & \oursspeed{\best{3.541}} & \oursspeed{\best{5.885}} & \oursspeed{\best{9.885}} & \oursspeed{\best{11.913}} & \oursspeed{\best{18.688}} & \oursspeed{\best{39.085}} & \oursspeed{\best{49.613}} \\
\midrule
\rowcolor{tableoneband}\multicolumn{11}{c}{\textcolor{tableoneblue}{\itshape\tableoneTokens}} \\
\midrule
\modelname{MemAgent}{7B} & \ratioval{\runnerup{1}} & \ratioval{\runnerup{1}} & \ratioval{\runnerup{1}} & \ratioval{\runnerup{1}} & \ratioval{1} & \ratioval{1} & \ratioval{1} & \ratioval{1} & \ratioval{1} & \ratioval{1} \\
\modelname{InfMem}{4B} & \ratioval{5.6} & \ratioval{4.4} & \ratioval{2.8} & \ratioval{1.2} & \ratioval{\runnerup{0.63}} & \ratioval{\runnerup{0.45}} & \ratioval{\runnerup{0.27}} & \ratioval{\runnerup{0.081}} & \ratioval{\runnerup{0.040}} & \ratioval{\runnerup{0.069}} \\
\modelname{Lychee ($-G$)}{7B} & \ratioval{1.99} & \ratioval{1.86} & \ratioval{1.54} & \ratioval{1.21} & \ratioval{0.92} & \ratioval{0.81} & \ratioval{0.92} & \ratioval{0.70} & \ratioval{0.89} & \ratioval{0.90} \\
\modelname{ReMemR1}{7B} & \ratioval{4.75} & \ratioval{4.40} & \ratioval{3.62} & \ratioval{2.82} & \ratioval{2.29} & \ratioval{2.11} & \ratioval{1.78} & \ratioval{1.57} & \ratioval{1.62} & \ratioval{1.28} \\
\rowcolor{tableoneours}\textcolor{tableoneblue}{\modelname{\textbf{MemDuet}\tableoneMemDuetMarker}{7B}} & \tableoneMemDuetTokenCells \\
\arrayrulecolor{tableoneoutline}\bottomrule
\end{tabularx}
\endgroup

\caption{HQA length scaling: accuracy, inference speedup, and relative output tokens.
Model sizes beside method names refer to the updater backbone; MemDuet is labeled by its 7B updater.
Qwen2.5-1M denotes Qwen2.5-Instruct-1M, DS-Distill denotes DS-Distill-Qwen,
and RAG and Search-R1 use Qwen2.5-7B-Instruct.
Lychee denotes LycheeMemory; $-G$/$+G$ indicate without/with its native gate.
Local accuracy is sub-EM (\%); paper rows retain their reported metric.
Paper-priority cells use
\href{https://arxiv.org/pdf/2507.02259v2}{MemAgent v2, Table 1} and
\href{https://aclanthology.org/2026.acl-long.365.pdf}{LycheeMemory, Table 1}
without local rescoring. RAG, Search-R1, and Lychee ($-G$) are unreported at
3.5M. ReMemR1 uses local scores because
its paper's MemAgent baseline differs from MemAgent v1; GRU-Mem also has a
different paper baseline but lacks matching local accuracy and is blank.
InfMem and MemDuet use local scores. Dashes mark unreported results; N/A
retains the unavailable entries for six full-context baselines at 1.75M and
3.5M (their 1.75M runs exceeded memory).
The middle and lower panels report time speedup and relative updater output
tokens separately. Their values are
$t_{\mathrm{MemAgent}}/t_{\mathrm{method}}$ and
$o_{\mathrm{method}}/o_{\mathrm{MemAgent}}$, respectively, using mean
per-question time $t$ and updater output tokens $o$. Predictor output is separate.
MemAgent is $1\times$ for both time speedup and relative output tokens;
lower token ratios mean fewer output tokens. GRU-Mem and Lychee ($+G$) are
omitted from the token panel because their comparable output-token counts
are unavailable. Within each length column and panel, the best value is
bold and the second-best distinct value is underlined; ties share a rank.
Missing entries are excluded. Rankings are descriptive and include
estimated speedups.
Values are rounded for display.
Paper-reported accuracy and local efficiency are not matched runs.
Time is normalized to A100-equivalent seconds using one paired 128-example
MemDuet workload; cross-hardware transfers are estimates, not hardware-neutral
measurements. GRU-Mem speedups are approximate readings of the paper's heatmap
against its own MemAgent baseline; Lychee ($+G$) speedups transfer reported
gated/ungated ratios onto local ungated timings, so neither is a matched local
measurement. $^{\dagger}$ denotes these paper-derived speedups.
$^{\ddagger}$The gray row is the historical V12 variant with an original 7B updater and a
frozen 2B predictor, not a joint-trained three-stage model. Its 3.5M speedup
is 49.61$\times$ and its relative output-token use is 0.00746$\times$ on four paired examples;
its 3.5M accuracy is from a separate 128-example aggregate and must not be
interpreted as the accuracy of those four paired runs.
Other long-context sample counts also vary. Dashes mean no numerical result was reported;
historical hardware and evaluation settings are not fully matched.}
\label{tab:discussion-length-scaling}
\end{table*}

%% file: sections/appendix_s1_backbone_results.tex
\subsection{\apptext{System 1 Backbone Comparison}}
\label{app:s1-backbone-results}

\begin{table*}[!t]
\centering
\begingroup
\definecolor{soneband}{HTML}{F1F3F7}
\definecolor{soneshade}{HTML}{EAF1FA}
\fontsize{7.8}{9.5}\selectfont
\setlength{\tabcolsep}{0.6pt}
\renewcommand{\arraystretch}{1.18}
\begin{tabularx}{\textwidth}{@{}l*{11}{>{\centering\arraybackslash}X}@{}}
\toprule
System~1 & 7K & 14K & 28K & 56K & 112K & 224K & 448K & 896K & 1.75M & 3.5M
  & \apptext{Mean}\\
\csname @@input\endcsname sections/s1_backbone_per_length_rows.tex
\bottomrule
\end{tabularx}
\endgroup
\caption{\apptext{Complete System~1 backbone results on RULER-HQA
(7K--3.5M tokens). Mean is the arithmetic average of the ten per-length
values. Speed and Tokens are ratios relative to RL-MemAgent.}}
\label{tab:s1-backbone-full}
\end{table*}

\apptext{Table~\ref{tab:s1-backbone-full} gives the complete per-length
results underlying Table~\ref{tab:s1_choice}. Accuracy, Speed, and Tokens
equally weight the ten context lengths from 7K to 3.5M tokens.
For each length, Speed is the ratio of mean RL-MemAgent latency to mean
MemDuet latency, and Tokens is the ratio of mean MemDuet output-token
count to mean RL-MemAgent output-token count.}

\apptext{}

%% file: sections/appendix_prompts_en.tex
\section{Prompt Templates}
\label{app:prompts}

The following prompts are used in MemDuet's training and inference,
covering candidate selection and verification, memory initialization and
updating, and final answer generation. The placeholders
\texttt{\{prompt\}}, \texttt{\{memory\}}, and \texttt{\{chunk\}}
denote the question, current memory, and input document chunk, respectively.

\subsection{Candidate Selection}
\label{app:prompt-gate}

Use: System~1 inference and memory-conditioned action supervision.
Inputs: question, current memory, and one raw chunk.
Output: one token, \texttt{k} (SKIP) or \texttt{u} (UPDATE), without a
memory draft or explanation. Parallel calls substitute the same committed
memory. For before/after supervised pairs, the memory placeholder changes
and the target action changes; the task text need not be duplicated.

\memorypromptpanel{Complete candidate-selection task text}{prompts/gate.txt}

\subsection{Candidate Verification}
\label{app:prompt-verification}

Use: System~2 verifies whether a selected chunk warrants a memory update
before memory generation.
Inputs and task text are identical to Appendix~\ref{app:prompt-gate},
with System~2's own chat template. Output interface: first-position
\texttt{k}/\texttt{u} and both action scores, consumed by the controller
in Appendix~\ref{app:inference-configuration} for open-weight inference.
The proprietary API adaptation instead parses generated \texttt{k}/\texttt{u}
labels without action scores, as described in Appendix~\ref{app:proprietary-api}.
There is no separate
natural-language \emph{accept/reject} prompt or small-model memory draft.

\subsection{Memory Initialization and Updating}
\label{app:prompt-memory-update}

Use: first-chunk initialization and committed rewrites.
Inputs: question, raw chunk, and previous memory, initialized with
\texttt{No previous memory}. Output: complete replacement memory, not
an append-only patch. The local MemAgent full-update baseline shares this
template; controller protocol wrappers are not extra model instructions.

\memorypromptpanel{Complete memory-initialization/update task text}{prompts/memory_update.txt}

\subsection{Final Answer Generation}
\label{app:prompt-final-answer}

Use: answering after document processing. Inputs: question and final
committed memory, without raw documents or discarded candidates.
Output: the answer in \texttt{\textbackslash boxed\{\}}, as below.

\memorypromptpanel{Complete final-answer task text}{prompts/final_answer.txt}

\subsection{LongBench v2 Multiple-Choice Prompt}
\label{app:prompt-longbench-v2}

For LongBench v2, the \texttt{\{prompt\}} placeholder in the preceding
templates contains the question, all four options, and the letter-answer
instruction shown in the \texttt{<problem>} block below. Candidate
selection, verification, and memory updates use this same question text.
The final-answer template combines it with the final memory and asks for
an option letter in \texttt{\textbackslash boxed\{\}}. Option accuracy is
defined in Appendix~\ref{app:metrics-protocol}.

\memorypromptpanel{Complete LongBench v2 final-answer task text}{prompts/longbench_v2_final_answer.txt}

%% file: sections/appendix_cases_en.tex
\section{Case Studies}
\label{app:cases}

We present three successful trajectories and one failure.
Evidence and memory excerpts
below are condensed from recorded states, not verbatim model transcripts.
Unshown chunks and intermediate rounds are explicitly omitted.
Here $d$ and $v$ denote the System~1 and System~2 UPDATE probabilities,
respectively; $\alpha$ is the acceptance probability and $\xi$ is the
sampled uniform variate.

\subsection{Successful Reasoning and Memory Evolution}
\label{app:case-bridge}

\begin{memorycase}{Successful case: completing a two-hop evidence chain}
Question: When was the British author who wrote the novel on which
\emph{Here We Go Round the Mulberry Bush} was based born?
\par\smallskip
Sample: \texttt{eval200}, ID 112; six chunks.
Key evidence: Document 109 identifies the novelist as Hunter Davies;
Document 177 gives his birth date, 7 January 1936.
\caseboundary
\casestep{Initialization and current memory}{
Chunk $c_1$ produces the initial memory: Clive Donner is mentioned as
the film director, while the novelist's identity and birth date remain
unresolved. The mandatory initialization is a System~2 memory call.}
\casestep{Chunk $c_3$ / Document 109}{
Evidence: The film is based on a novel by Hunter Davies.\par
Current memory: The target author's identity is unresolved.\par
System~1: Select UPDATE, $d=0.99903$.\par
System~2: $v=0.97069$, $\alpha=0.97163$, $\xi=0.43525$; accept.\par
Updated memory: Novelist = Hunter Davies; birth date unresolved.}
\casestep{Chunk $c_5$ / Document 177}{
Evidence: Edward Hunter Davies was born on 7 January 1936.\par
Current memory: Hunter Davies is the novelist.\par
System~1: Reevaluate after the $c_3$ commit and select UPDATE,
$d=0.99983$.\par
System~2: $v=0.99850$, $\alpha=0.99867$, $\xi=0.27546$; accept.\par
Updated memory: Hunter Davies is the novelist and was born on
7 January 1936.}
\casestep{Remaining processing and answer}{
The original $c_5$ prediction under the first snapshot was provisionally
accepted but discarded after the earlier $c_3$ commit; it did not generate
a memory. Direct SKIP decisions at $c_2$, $c_4$, and $c_6$ do not call the
verifier. Distractor text and other window details are omitted.\par
Final answer: \casecorrect{7 January 1936} (matches the reference).\par
Memory generations: three, including initialization.}
\end{memorycase}

The trajectory joins the film--novelist fact in $c_3$ with the birth-date
fact in $c_5$. The small model's output is a chunk-selection action; all
displayed memories were generated by System~2. Provisional acceptance and
commitment in document order are separate events.

\begin{memorycase}{Successful case: changing selection after a memory update}
Question: According to the 2001 census, what was the population of the
city in which Kirton End is located?
\par\smallskip
Sample: \texttt{eval200}, ID 58; six chunks.
Key evidence: Document 81 places Kirton End in the Boston district;
Document 96 gives Boston town's 2001 population as 35,124.
\caseboundary
\casestep{Initialization and current memory}{
Document 10 in $c_1$ describes Kirton, Suffolk, with a 2011 population of
1,146. The initial memory retains that information but explicitly leaves
the requested city's 2001 population unresolved.}
\casestep{Chunk $c_3$ / Documents 81 and 96}{
Evidence: Kirton End is in the Boston district of Lincolnshire. Boston
town had 35,124 residents in 2001, whereas the borough's 2015 population
was 66,900.\par
Current memory: The location--population chain is incomplete.\par
System~1: Select UPDATE, $d=0.99984$.\par
System~2: $v=0.88080$, $\alpha=0.88094$, $\xi=0.28715$; accept.\par
Updated memory: Links Kirton End to Boston and records 35,124 as the
town's population for the requested census year.}
\casestep{Same chunk $c_4$, before and after commitment}{
Under the earlier memory snapshot, $c_4$ is selected as UPDATE
($d=0.09009$, $v=0.09535$, $\alpha=1$). This provisional decision is
discarded when the earlier $c_3$ update is committed.\par
Under the new memory, System~1 reevaluates $c_4$ as SKIP
($d=0.000970$); System~2 is not called for this decision and the memory
remains unchanged. The chunk text is unchanged; its full text is omitted.}
\casestep{Remaining processing and answer}{
Chunks $c_2$, $c_5$, and $c_6$ are also skipped in their effective
decisions. Other window details are omitted.\par
Final answer: \casecorrect{35,124} (matches the reference).\par
Memory generations: two, at $c_1$ and $c_3$.}
\end{memorycase}

The committed memory distinguishes the town from the borough and the
2001 census from later population estimates. The same $c_4$ changes from
UPDATE to SKIP after that rewrite, directly illustrating memory-conditioned
selection and the invalidation of decisions made under an earlier state.
This is an observed trajectory, not a counterfactual intervention.

\begin{memorycase}{Successful case: maintaining sufficient memory}
Question: Hayden is a singer-songwriter from Canada, but where does
Buck-Tick hail from?
\par\smallskip
Sample: \texttt{eval800}, ID 31; 23 chunks.
Key evidence: Document 1 states that Buck-Tick was formed in Fujioka,
Gunma, in 1983.
\caseboundary
\casestep{Initialization and current memory}{
Chunk $c_1$ contains Document 1. System~2 initializes the memory with the
band's origin, Fujioka, Gunma, already supplying the requested answer.}
\casestep{Later candidate at $c_9$}{
Current memory: Buck-Tick hails from Fujioka, Gunma.\par
System~1: Select UPDATE, $d=0.99948$.\par
System~2: $v=0.56218$, $\alpha=0.56247$, $\xi=0.77329$;
reject because $\xi>\alpha$.\par
Memory after verification: Unchanged; no memory rewrite is generated.
The candidate's full text is omitted.}
\casestep{Other chunks and final answer}{
The other 21 chunks after initialization receive direct SKIP decisions
and do not call System~2. The final memory is identical to the initial
memory.\par
Final answer: \casecorrect{Fujioka, Gunma} (matches the reference).\par
Memory generations: one, for initialization only.}
\end{memorycase}

This case shows a different successful path: the necessary evidence is
available immediately, and later processing preserves it without further
memory generation. The $c_9$ rejection is a sampled acceptance-rule outcome,
not evidence that its text is factually false or necessarily irrelevant.
The generation count describes this trajectory; it is not a latency
measurement or a general efficiency estimate.

\subsection{Failure Analysis}
\label{app:failure-cases}

\begin{failurecase}{Failure case: confusing the director with the novelist}
Question: When was the author of the novel behind
\emph{Here We Go Round the Mulberry Bush} born?
\par\smallskip
Sample: \texttt{eval800}, ID 112; 23 chunks.
Key evidence: Document 187 gives director Clive Donner's birth date;
Document 366 gives Hunter Davies's birth date; Document 751 identifies
Davies as the novelist.
\caseboundary
\failurestep{Current memory before $c_6$}{
The initial memory does not identify the target novelist and contains
unresolved, partly irrelevant biographical information. Intervening
uncommitted candidates and distractor text are omitted.}
\failurestep{Chunk $c_6$ / Document 187}{
Evidence: Clive Donner, a film director, was born on 21 January 1926.\par
System~1: Select UPDATE, $d=0.20181$.\par
System~2: $v=0.97702$, $\alpha=1$, $\xi=0.52394$; accept.\par
Updated memory: \casewrong{Clive Donner is treated as the novel's author},
and his birth date is attached to the target. A later $c_7$ rewrite retains
this incorrect role binding; its full text is omitted.}
\failurestep{Earlier supporting attribute / Document 366}{
Evidence: Hunter Davies was born on 7 January 1936.\par
Recorded outcome: This date never appears in a committed memory.
The trace does not independently label which gate decision caused this
omission, so it is not counted as a verified false SKIP.}
\failurestep{Chunk $c_{22}$ / Document 751}{
Current memory: The incorrect Donner--author association and date persist.\par
Evidence: The novelist is Hunter Davies.\par
System~1: Select UPDATE, $d=0.11920$.\par
System~2: $v=0.43782$, $\alpha=1$, $\xi=0.03303$; accept.\par
Updated memory: Identifies Davies but lacks his birth date, retains
Donner's date, and adds an unsupported ordering of their ages.}
\failurestep{Final answer}{
Prediction: \casewrong{21 January 1926}.\par
Reference: \casecorrect{7 January 1936}.\par
Memory generations: four, including initialization. Other rounds are omitted.}
\end{failurecase}

The $c_6$ rewrite binds the director's birth date to the novelist; later
evidence repairs the name but not the date. This reveals incomplete memory
repair. Acceptance does not certify truth, and the trace establishes neither
a causal false SKIP nor failure prevalence. The shorter successful case
differs in evidence order and distractors, precluding a controlled length
comparison.